\documentclass[final,3p,times]{elsarticle}

\AtBeginDocument{%
  \def\thefnote{\myfnsymbol{fnote}}}

\makeatletter
\def\myfnsymbol#1{\expandafter\@myfnsymbol\csname c@#1\endcsname}
\def\@myfnsymbol#1{%
  \ifcase #1
    \or $\dagger$%
    \or $\ddagger$%
    \else \@ctrerr%
  \fi}
\def\fntext[#1]#2{%
  \g@addto@macro\@fnotes{%
    \refstepcounter{fnote}\elsLabel{#1}%
    \def\thefootnote{\thefnote}%
    \global\setcounter{footnote}{\c@fnote}%
    \footnotetext{#2}}}
\makeatother

\usepackage{amssymb}
\usepackage{amsmath}

\usepackage{booktabs}
\usepackage{graphicx}
\usepackage{longtable}
\usepackage{verbatim}
\usepackage{amsmath,amsfonts}
\usepackage[dvipsnames]{xcolor}
\usepackage{hyperref}
\usepackage{subcaption}
\usepackage[htt]{hyphenat}

\begin{document}

\begin{frontmatter}



\title{\includegraphics[height=1em, trim=4 50 4 4, clip]{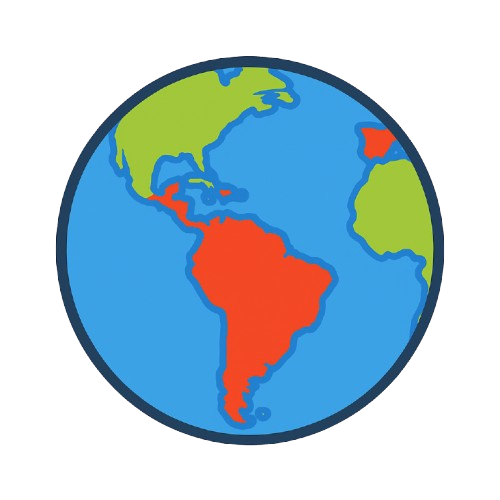} IberBench: LLM Evaluation on Iberian Languages}

 \author[label1]{José Ángel González}
 \affiliation[label1]{organization={Symanto Research},
             city={Valencia},
             postcode={46011},
            country={Spain}}
 \ead{jose.gonzalez@symanto.com}

\author[label1]{Ian Borrego Obrador}
\ead{ian.borrego@symanto.com}
\author[label2]{Álvaro Romo Herrero}
 \affiliation[label2]{
  organization={Keepler Data Tech},
  city={Madrid},
  postcode={28014},
  country={Spain}
 }
\ead{alvaro.romo@keepler.io}
\author[label1]{Areg Mikael Sarvazyan}
\ead{areg.sarvazyan@symanto.com}
\author[label1]{Mara Chinea-Ríos}
\ead{mara.chinea@symanto.com}
\author[label3]{Angelo Basile}
 \affiliation[label3]{
  organization={Universitat Politècnica de València},
  city={Valencia},
  postcode={46022},
  country={Spain}
 }
\ead{abasile@doctor.upv.es}
\author[label4]{Marc Franco-Salvador \fnref{unicc_disclaimer}}
 \affiliation[label4]{
  organization={United Nations International Computing Centre (UNICC)},
  city={Valencia},
  postcode={46930},
  country={Spain}
 }
\ead{francom@unicc.org}
\fntext[unicc_disclaimer]{This publication does not reflect the position/views of UNICC, and only the authors’ views.}

\begin{abstract}
Despite their remarkable success, Large Language Models (LLMs) remain difficult to evaluate comprehensively, particularly for languages other than English, where high-quality data is often limited. Existing benchmarks and leaderboards are predominantly English-centric, with only a few addressing other languages. These benchmarks fall short in several key areas: they overlook the diversity of language varieties, prioritize fundamental Natural Language Processing (NLP) capabilities over tasks of industrial relevance, and are static. With these aspects in mind, we present IberBench, a comprehensive and extensible benchmark designed to assess LLM performance on both fundamental and industry-relevant NLP tasks, in languages spoken across the Iberian Peninsula and Ibero-America, including Spanish, Portuguese, Catalan, Basque, Galician, and English, along with Spanish varieties like Mexican, Uruguayan, Peruvian, Costa Rican, and Cuban.
IberBench integrates 101 datasets from evaluation campaigns and recent benchmarks, covering 22 task categories such as sentiment and emotion analysis, toxicity detection, and summarization. 
The benchmark addresses key limitations in current evaluation practices, such as the lack of linguistic diversity and static evaluation setups by enabling continual updates and community-driven model and dataset submissions moderated by a committee of experts.
We evaluate 23 LLMs ranging from 100 million to 14 billion parameters and provide empirical insights into their strengths and limitations. Our findings indicate that (i) LLMs perform worse on industry-relevant tasks than in fundamental ones, (ii) performance is on average lower for Galician and Basque, (iii) some tasks show results close to random, and (iv) in other tasks LLMs perform above random but below shared task systems.
IberBench offers open-source implementations for the entire evaluation pipeline, including dataset normalization and hosting, incremental evaluation of LLMs, and a publicly accessible leaderboard.\footnote{The Leaderboard UI is accessible at \url{https://huggingface.co/spaces/iberbench/leaderboard}}
\end{abstract}








\end{frontmatter}




\section{Introduction} \label{sec:introduction}
LLMs have revolutionized NLP since they understand and generate language in zero- and few-shot settings, they are flexible in handling almost any task, they are easy to use, and they show strong multilingual capabilities. LLM adoption is growing rapidly among individuals and organizations, to the point that LLMs are used to automate everyday tasks like writing or editing emails, and core industrial services such as sentiment analysis for market insights \cite{eloundou2023gptsgptsearlylook, 10500411}.

Evaluating LLMs is essential to understand their strengths in such tasks and promote their adoption. Often, we are addressing a task $\mathcal{T}$ with LLM $\mathcal{A}$, and we want to know whether LLM $\mathcal{B}$ is better than $\mathcal{A}$ in $\mathcal{T}$. Many times, $\mathcal{A}$ is selected by \textit{a priori} decision, but if we are able to determine that $\mathcal{B}$ is better for $\mathcal{T}$, $\mathcal{B}$ will be adopted instead. There are many LLMs available, generally falling into two categories: open-source models such as Llama \cite{grattafiori2024llama3herdmodels}, Qwen \cite{qwen2025qwen25technicalreport}, and DeepSeek \cite{deepseekai2025deepseekr1incentivizingreasoningcapability}, and closed-source models such as GPT \cite{openai2024gpt4technicalreport} and Gemini \cite{geminiteam2024gemini15unlockingmultimodal}. The landscape is vast, with variations in reasoning-focused and chat-based models, monolingual and multilingual capabilities, parameter size, fine-tuning strategy, and more. This diversity makes choosing the right LLM for a specific task and language challenging, as factors such as performance, latency, and cost must be carefully considered.

Extensive research has been conducted on LLM evaluation. There are, at least, two main approaches: (i) automatic evaluation using carefully designed benchmarks and (ii) manual evaluation, where real users interact with models and express their preferences. Notable examples of the former include HELM \cite{liang2023holistic} and the Open LLM Leaderboard \cite{open-llm-leaderboard-v2}, while Chatbot Arena \cite{10.5555/3692070.3692401} is the most well-known example of the latter. However, most benchmarks following these approaches lack at least one of these dimensions: multilingual coverage, industrial interest, and immutability.

Most evaluations of LLM capabilities prioritize English, often overlooking the rich linguistic diversity of other languages. This gap is particularly evident when it comes to the languages of the Iberian Peninsula—Spanish, Portuguese, Catalan, Basque, and Galician—as well as the many regional varieties of Spanish and Portuguese spoken across Ibero-America, such as Mexican, Peruvian, Cuban, or Uruguayan.\footnote{For simplicity, we will refer to all these languages collectively as \textit{Iberian languages}.} Despite having over 800 million speakers, Iberian languages remain underrepresented in LLM evaluations, with some recent attempts to alleviate this \cite{laleaderboard2024, baucells-etal-2025-iberobench, odesia2024}. Most of these benchmarks focus on fundamental language capabilities and knowledge-based tasks, such as question answering, language understanding, proficiency, and textual entailment. They overlook industry-relevant tasks like intent classification, toxicity detection, affective analysis, and user profiling. Another major limitation of existing benchmarks is their static nature: once established, the datasets remain unchanged over time. This lack of adaptability increases the risk of saturation and prevents the inclusion of novel tasks and languages, ultimately limiting their long-term utility.

\begin{figure}[h]
    \centering
    \begin{subfigure}[b]{0.49\textwidth}
        \centering
        \includegraphics[width=\textwidth, height=5cm]{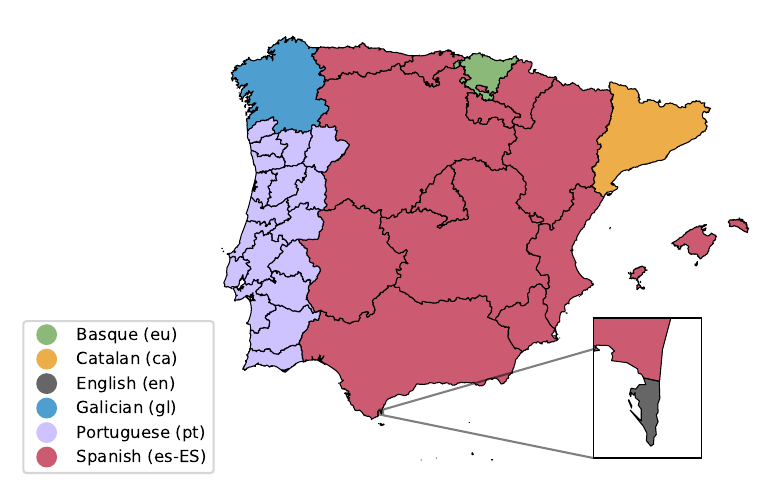}
        \caption{Iberian Peninsula languages}
    \end{subfigure}
    \hfill
    \begin{subfigure}[b]{0.49\textwidth}
        \centering
        \includegraphics[width=0.7\textwidth, height=5cm]{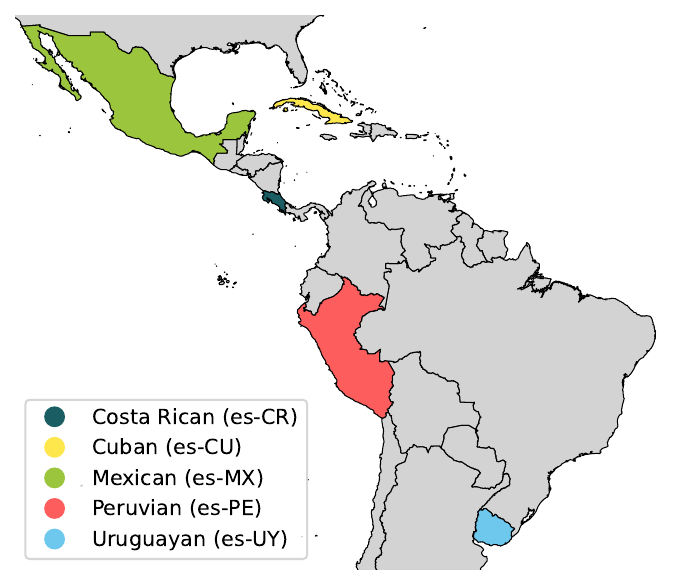}
        \caption{Spanish varieties in Ibero-America}
    \end{subfigure}
    \caption{\label{map-iberian-languages}Maps showing the Iberian languages considered in IberBench.}
    \label{fig:language_maps}
\end{figure}

To fill these gaps, we introduce IberBench, a benchmark specifically crafted to continually evaluate LLMs across both fundamental NLP tasks and industry-relevant challenges in Iberian languages. IberBench draws its datasets from two main sources. The first is a collection of shared tasks from workshops like IberLEF \cite{iberlef2024}, IberEval \cite{ibereval}, TASS \cite{DBLP:conf/sepln/2018tass}, and PAN \cite{DBLP:conf/clef/StamatatosPPRS15}, spanning editions from 2014 to 2024, co-located with the Spanish Society for Natural Language Processing (SEPLN) and the Conference and Labs of the Evaluation Forum (CLEF). These workshops have played a key role in advancing research and fostering collaboration within the Iberian NLP community, providing a wealth of practical, industry-aligned tasks. The second source consists of recent general-purpose benchmarks specifically designed to assess LLMs on fundamental language tasks. This includes near all datasets from La Leaderboard \cite{laleaderboard2024}, the evaluation suite of the Latxa model \cite{etxaniz-etal-2024-latxa}, and others. IberBench brings together a diverse collection of datasets developed over the years for training and evaluating NLP models, combining both workshop-based resources and contributions from the broader scientific literature. In total, IberBench comprises 101 datasets, tailored to the Iberian languages shown in Figure \ref{map-iberian-languages} and spanning 22 task categories.

Beyond the datasets, three key components orchestrate the evaluation of LLMs in IberBench.
First, the Leaderboard UI serves as the main interface, displaying various LLM rankings, plots, and reports; allowing users to request new models for evaluation; and describing the datasets included in the benchmark.
The second component is the introduction of a committee of NLP specialists from both academia and industry which is responsible for key decisions, such as determining which user-requested LLMs should be evaluated and which new datasets should be included. 
Finally, the third component is the evaluation framework. For this, we rely on \texttt{lm-evaluation-harness} \cite{eval-harness}, which we have extended to (i) include the IberBench datasets, (ii) support sequence labeling tasks, and (iii) streamline the evaluation process by integrating a caching mechanism to add new datasets and evaluate LLMs solely on this data. This way we ensure the relevance and applicability of IberBench over time.

So far, we have evaluated 23 LLMs, ranging from 100 million to 14 billion parameters. These include major multilingual families like Llama, Qwen, and Phi \cite{abdin2024phi4technicalreport}, as well as models tailored for Iberian and European languages, such as Latxa \cite{etxaniz-etal-2024-latxa}, Salamandra \cite{gonzalezagirre2025salamandratechnicalreport}, and EuroLLM \cite{MARTINS202553}. Our main findings reveal that (i) LLMs struggle with industry-relevant tasks compared to fundamental language tasks, which dominate existing benchmarks, (ii) Galician and Basque present greater challenges than other languages, (iii) some tasks like lexical borrowing detection, intent classification, and machine-generated text detection remain largely unsolved, with top-performing LLMs barely surpassing a random guesser, and (iv) in other tasks such as sentiment analysis, humor detection, and fake news detection, LLMs are better than the random baseline but worse than most shared task submissions.

In summary, these are our key contributions:
\begin{itemize}
\item We introduce IberBench, a comprehensive benchmark comprising 101 datasets that span a wide range of fundamental and industry-relevant NLP tasks, broadening the scope of LLM evaluations in Iberian languages. The IberBench pipeline is publicly released.

\item We design IberBench with scalability, extensibility, and reproducibility in mind, to seamlessly integrate new datasets, language varieties, and models over time, ensuring its continued relevance as the field evolves.

\item We collect and standardize 58 datasets from workshops on Iberian languages—many of which are difficult to access—and make them readily available through a unified evaluation framework.

\item We evaluate both multilingual and language-specific models, providing insights into their performance on tasks that reflect the linguistic diversity of Iberian languages.

\end{itemize}
\section{Related Work} \label{sec:related_work}

LLM evaluation schemes can often be categorized into two perspectives: human and automatic. 
Human evaluation is subjective, often to measure specific behavioral properties or human preferences to model outputs.
It estimates output quality by crafting tailored inputs to elicit specific outputs from LLMs and assess user-centric or ethical aspects such as helpfulness, fairness, security or safety.
Prominent examples include evaluating a model's robustness to red-teaming and jailbreaking attempts \cite{xu-etal-2021-bot, wei2023jailbroken}, multi-turn conversation and instruction following abilities \cite{bai-etal-2024-mt}, and general human preference toward a specific model's outputs \cite{10.5555/3692070.3692401, li2024crowdsourceddatahighqualitybenchmarks}.
Unfortunately, human evaluation of LLMs incurs high costs and is often biased, making it inadequate for large-scale, objective, reproducible model evaluations \cite{wu-aji-2025-style, schoch-etal-2020-problem}.
Moreover, human evaluation can include malicious actors that may poison the results \cite{baumann2024universal}.

Automatic evaluation is faster, more cost effective, less bias-prone and reproducible \cite{eval-harness}.
It is often used to measure performance in industry-relevant tasks like sentiment analysis or topic modeling, and fundamental capabilities like natural language understanding, generation, and question answering. 
Automatic benchmarks are often created by collecting existing datasets, releasing them along with LLM leaderboards and automatic evaluation pipelines.
Some of the first popular benchmarks include GLUE \cite{wang-etal-2018-glue} and MMLU \cite{hendryckstest2021}, which mainly measured fundamental capabilities like reading comprehension, common sense reasoning, or general knowledge, while more recent efforts like LM Eval Harness \cite{eval-harness}, HELM \cite{liang2023holistic}, and the Open LLM Leaderboard \cite{open-llm-leaderboard-v2} also include a range of industry-relevant tasks such as domain-specific sentiment analysis, social media toxicity detection, programming, and mathematical reasoning.
Moreover, recent advances in LLM-as-a-judge techniques show that LLMs are good alternatives to humans in specific scenarios \cite{DBLP:journals/corr/abs-2406-18403} when evaluating behavioral properties. It is proven that LLMs generate highly correlated answers to humans \cite{10.5555/3666122.3668142}, and exhibiting similar bias levels \cite{chen-etal-2024-humans}. 

While most of the work has been done for English capability evaluations, there are recent efforts focusing on non-English languages, with new benchmarks for European \cite{ali2024teukenn}, Indic \cite{singh-etal-2024-indicgenbench}, South-east Asian \cite{susanto2025seahelmsoutheastasianholistic}, Arabic \cite{almazrouei-etal-2023-alghafa} languages, and more. 
Spanish co-official languages have been represented in various initiatives such as La Leaderboard \cite{laleaderboard2024}, Iber\underline{o}Bench \cite{baucells-etal-2025-iberobench},\footnote{Despite the name similarity, IberBench and Iber\underline{o}Bench were developed independently and in parallel, without the involved parties being aware of each other.} and Odesia \cite{odesia2024}. However, these efforts still fail to capture the full extent of linguistic variation in Iberian languages, they primarily focus on fundamental language and knowledge tasks, and are static. La Leaderboard and Iber\underline{o}Bench focus on language understanding datasets sourced from multilingual human translations of existing English corpora, while Odesia focuses on a small set of sexism and propaganda detection datasets from existing shared tasks in the IberLEF evaluation campaigns.
IberBench extends these approaches to include (i) both fundamental and industry-relevant tasks from workshops like IberLEF, IberEval, TASS, and PAN, (ii) not only Spain's co-official languages but also language varieties from Ibero-America, and (iii) a scalable framework to extend the benchmark with new datasets and language varieties over time.

Most existing benchmarks focus on classification or generation tasks, largely ignoring sequence labeling tasks such as Named Entity Recognition (NER) and Aspect-Based Sentiment Analysis (ABSA), that are essential for industrial applications.
Recent studies indicate that LLMs can match or even surpass encoder models in sequence labeling tasks \cite{dukic-snajder-2024-looking, wang2023gptnernamedentityrecognition}, however, since current benchmarks largely overlook these tasks, LLMs have yet to be thoroughly evaluated in this scenario. Bearing this in mind, IberBench allows to evaluate LLMs in sequence-labeling tasks through a custom evaluation scheme, integrated in \texttt{lm-evaluation-harness}.
\section{Iberbench} \label{sec:iberbench}

\begin{figure}[t]
\centering
\includegraphics[scale=0.08]{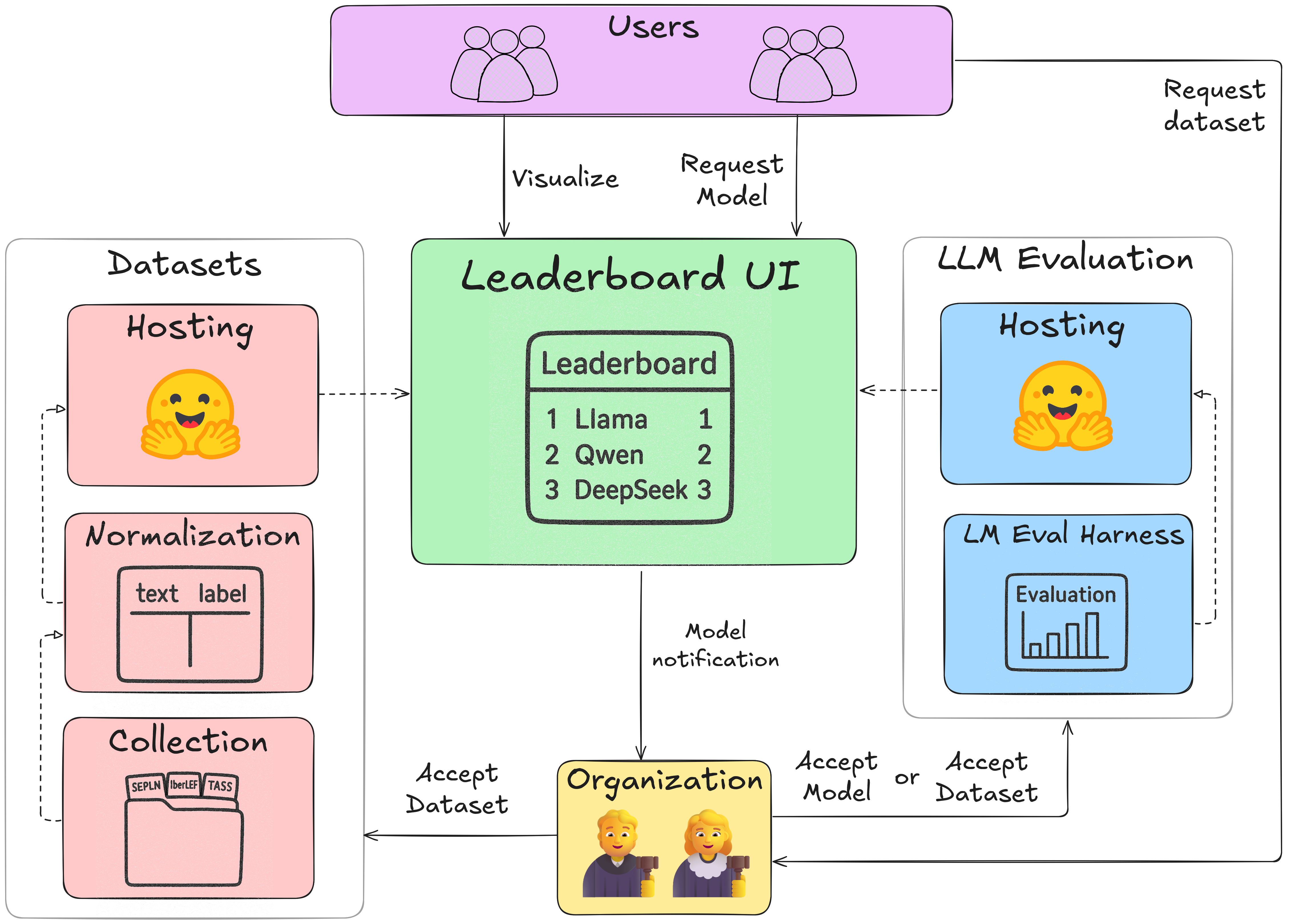}
\caption{\label{fig:iberbench-diagram} IberBench overview. Users can view rankings, plots, and reports; request LLMs for evaluation through the UI; and propose new datasets to the organization. The organization reviews these proposals for possible inclusion in the leaderboard. Once approved, datasets and models are prepared, evaluated, and hosted to be displayed in the UI.}
\end{figure}

IberBench is made up of four key components that manage interactions, datasets, and evaluations: leaderboard UI, organization, datasets, and LLM evaluation. The entire pipeline of IberBench is illustrated in Figure~\ref{fig:iberbench-diagram}.

The benchmark’s entry point is the leaderboard UI, hosted on a HuggingFace Space,\footnote{The Leaderboard UI is accessible at \url{https://huggingface.co/spaces/iberbench/leaderboard}} where users can view rankings, plots and detailed reports of the evaluation results, submit new models for evaluation, and explore the datasets of the benchmark. When a new LLM is requested, the request is queued to be reviewed by the IberBench organization, which decides whether to include it in the leaderboard.\footnote{The organization space is accessible at \url{https://huggingface.co/iberbench}} In the same way, users can request the addition of new datasets through discussions with the organization.\footnote{The discussions space is accessible at \url{https://huggingface.co/spaces/iberbench/README/discussions}} Upon acceptance, the new datasets will be downloaded, normalized, and hosted in a private HuggingFace repository. Each time a new model or dataset is accepted and prepared, the organization will run the LLM Evaluation through a custom evaluation module that relies on \texttt{lm-evaluation-harness},\footnote{The evaluation module is accessible at \url{https://github.com/IberBench/iberbench-evaluation/}} and the results will be reflected on the leaderboard UI. All these components are described in more detail in sections below.

\subsection{Leaderboard UI}
Inspired by prominent existing benchmarks \cite{open-llm-leaderboard-v2, ali2024teukenn}, IberBench introduces a comprehensive, interactive, and user-friendly leaderboard interface, hosted in a dedicated HuggingFace Space, that presents model rankings across task types, languages, and language varieties. The interface includes a wide array of plots, reports, and statistics, enabling users to perform fine-grained analyses and gain a detailed understanding of LLM capabilities for Iberian languages. Users can compare model performance across languages, domains, and tasks, facilitating both broad and targeted evaluations. The leaderboard displays general rankings across all Iberian languages, as well as language-specific rankings that highlight model performance on individual tasks within each language or variety. Each ranking includes detailed metadata such as model name, parameter size, and evaluation scores, aggregated by language, variety, and task type. The interface also provides a suite of visualizations, including plots to see what are the top-performing models on average, across tasks, Iberian languages, and performance trends with respect to model size, model type, and task type.

To support community-driven evaluations, the interface includes a request portal that guides users through the process of requesting new models. The request form requires: (i) the model or adapter name as listed on the HuggingFace Hub \cite{houlsby2019parameter}, (ii) a short model description, (iii) a contact email for correspondence, (iv) the weight precision format (e.g., bfloat16, 8-bit, or quantization methods such as GPTQ \cite{frantar2023optq}), (v) the weight type —original or adapted, (vi) the base model (only for adapted models), and (vii) whether the model is pre-trained, fine-tuned, or fine-tuned for chat applications.
These model details directly influence the evaluation process, as they determine how models are loaded and which inference parameters are applied in evaluation. Certain attributes such as the precision format, can significantly affect inference behavior and performance outcomes. For this reason, and to ensure the novelty, compatibility, and significance of the models, all requests are subject to a review process by the organization. Upon acceptance, the model undergoes a rigorous, consistent, and reproducible evaluation process through \texttt{lm-evaluation-harness}.

Users can also explore the descriptions of the datasets associated with each language in the benchmark through the visualization portal. This is essential to ensure transparency and to properly acknowledge the contributions of dataset creators. All datasets included in IberBench have been developed over the years by the NLP community dedicated to Iberian languages and it is therefore critical to give due credit. To this end, the leaderboard interface includes a dedicated section that includes dataset details such as the source, task type,language, variety, number of labels (if applicable), and URL of the creator's website. In addition, some of the rankings and visualizations group tasks into categories to enable broader insights. For example, aggressiveness and offensive language detection tasks are grouped under Toxicity and Harmful Language Detection, while tasks such as Text Summarization and Intent Classification are categorized as industry-relevant tasks. To maintain transparency in this aggregation process, the visualization portal explicitly displays the mappings used to define these groups.

\subsection{Organization}
The IberBench organization serves as the central entity responsible for maintaining the high-quality standards that ensure the long-term reliability and impact of the benchmark. At the time of writing, it is composed of seven members, including experts in NLP, engineering, language ethics, and gender bias, drawn from both academia and industry. Most of its members have previously served as shared-task participants, organizers, or members of committees in workshops like IberLEF and PAN, bringing valuable experience and domain-specific knowledge. Their collective expertise plays a key role in guiding the development of IberBench, ensuring its methodological rigor, ethical integrity, and continued relevance to the research community.

Each time a new model or dataset is proposed, or when a user requests to join the organization, the team assesses the request based on internal acceptance criteria. For model requests, priority is given to: (i) models trained exclusively on Iberian languages, (ii) multilingual models in which Iberian languages constitute more than 30\% of the training tokens, and (iii) models developed in Europe, particularly within the Iberian Peninsula or Ibero-America. For dataset inclusion, preference is given to datasets that: (i) focus specifically on Iberian languages, (ii) originate from workshops organized in the Iberian Peninsula or Ibero-America, and (iii) target domains, language varieties, or tasks not yet represented in the benchmark. The organization is also responsible for reviewing applications from prospective new members. In this regard, preference is given to individuals from both academia and industry with extensive experience working with Iberian languages, particularly those who have organized shared tasks, participated in evaluation campaigns, or contributed to the creation of datasets in this linguistic context.

The role of the IberBench organization also involves dataset collection and model evaluation. Whenever a model is accepted, designated members of the organization are responsible for running the evaluation pipeline with this model on the existing IberBench datasets, either as batch or individual jobs, by leveraging cached results. Upon the acceptance of a dataset, members are tasked with downloading, normalizing, and storing the dataset in a long-term repository on HuggingFace, privately hosted under the IberBench HuggingFace organization. Additionally, they must create a well-grounded configuration file for the \texttt{lm-evaluation-harness} to ensure that the evaluation process is conducted fairly and consistently. Detailed descriptions of these procedures are provided in the following sections.

\subsection{Datasets}
\label{sec:datasets}
Datasets are a core part of the IberBench benchmark, used to evaluate models across a wide range of NLP tasks. As the first step in building IberBench, we carefully collected and prepared a set of 101 existing datasets to make them suitable for LLM evaluation. This initial collection lays the groundwork for a future expansion with new datasets from upcoming evaluation campaigns. All datasets were sourced from existing workshops and benchmarks.

Many of the datasets included in IberBench are sourced from recurring workshops and evaluation campaigns such as IberLEF \cite{iberlef2024}, IberEVAL \cite{ibereval}, and TASS \cite{DBLP:conf/sepln/2018tass}—organized by SEPLN—as well as PAN \cite{rangel2017overview}, held within CLEF. These datasets are often scattered across different platforms and are not always easily accessible, typically requiring users to request access directly from the organizers through various channels. Due to these restrictions, we contacted the organizers of all shared tasks from these workshops to obtain explicit permission for inclusion in IberBench. We only incorporate into the benchmark those datasets for which we were granted authorization by the creators. This process resulted in the retrospective compilation of 58 workshop-sourced datasets from 2014 to 2024. The remainder of the IberBench datasets include the evaluation suite used to assess the Latxa model, the Belebele benchmark \cite{bandarkar-etal-2024-belebele} for multilingual natural language understanding, and nearly all datasets included in La Leaderboard. This integration aims to unify recent LLM benchmarks with curated workshop datasets into a single benchmark that spans a broad range of tasks, Iberian languages, and both foundational and industry-relevant task types.

During the dataset collection phase, we communicated IberBench’s data usage policy to all dataset creators. Specifically, IberBench adheres to best practices regarding data management and ethical reuse: (i) dataset creators are appropriately credited on the interface and dataset cards; (ii) we do not claim ownership of any dataset; and (iii) all datasets are maintained in private repositories to prevent leakage and potential contamination. Preventing contamination is particularly critical, as evaluation results become unreliable if an LLM has been exposed to test data during training \cite{schaeffer2023pretraining}. To this end, datasets are never released publicly and are only accessed during evaluation through \textit{internal, secure servers}. However, we acknowledge that our measures can only prevent contamination from IberBench’s infrastructure, and cannot account for prior public release of the datasets by their original authors.

Following the initial collection phase, all datasets were standardized using a custom normalization pipeline developed to automate this process.\footnote{The normalization pipeline is available at \url{https://github.com/IberBench/dataset-preparation}} The pipeline is designed to handle datasets in diverse formats, including Excel spreadsheets, tabular or comma-separated value files, HuggingFace datasets, and plaintext. The pipeline performs a series of processing steps: it merges multiple files when necessary (e.g., when sample identifiers and labels are stored separately), removes extraneous columns not relevant to the task, standardizes column names (e.g., to “text” and “label”), adds language and language variety annotations where applicable, and cleans textual artifacts such as mojibake and mixed encodings (e.g., ``café'' is recovered from ``cafÃ©'') . Once processed, the normalized datasets are uploaded to a private HuggingFace repository \cite{lhoest-etal-2021-datasets}, accompanied by metadata such as the originating workshop, year, and label set. 

While this process is relatively straightforward for text classification and generation tasks, sequence labeling tasks pose additional challenges. As they require a specific annotation schema, we designed one inspired by prior work~\cite{wang2023gptnernamedentityrecognition, Hu2024-jf}. In such cases, LLMs must be given explicit instructions on how to produce span-annotated outputs. To this end, we include annotated examples in the datasets to be used later as few-shot examples. We define an annotation schema where output labels are directly included into the input text by enclosing each annotated span within corresponding label tags. An example of dataset preparation and evaluation of a NER task using this schema is shown in Table \ref{tab:ner-example}. We prepare an annotated example for each text and its reference IOB labels in the dataset. During evaluation, some annotated examples from the dataset are used as shots to instruct the LLM to follow the schema. We then parse the generated output to extract the predicted sequence of IOB labels. Detailed instructions for preparing sequence labeling datasets within IberBench are available in the documentation.\footnote{We provide a guide to prepare sequence-labeling datasets at \url{https://github.com/IberBench/iberbench-evaluation/blob/main/docs/token_classification_guide.md}}

\begin{table}[t]
\small
\begin{tabular}{ll}
\toprule
\multicolumn{2}{c}{\textbf{Dataset preparation}} \\ \midrule
\textbf{Text}            & \begin{tabular}[c]{@{}l@{}}\parbox[t]{5in}{Real Madrid lose against Valencia CF in Madrid}\end{tabular} \\ \midrule
\textbf{Reference labels}  & \begin{tabular}[c]{@{}l@{}}\parbox[t]{5in}{[{\color{ForestGreen}B-football\_team}, {\color{ForestGreen}I-football\_team}, O, O, {\color{ForestGreen}B-football\_team}, {\color{ForestGreen}I-football\_team}, O, {\color{RubineRed}B-location}]}\end{tabular} \\ \midrule
\textbf{Annotated example} & \begin{tabular}[c]{@{}l@{}}\parbox[t]{5in}{\textless response\textgreater~\textless {\color{ForestGreen}football\_team}\textgreater~Real Madrid \textless /{\color{ForestGreen}football\_team}\textgreater~lose against \textless {\color{ForestGreen}football\_team}\textgreater~Valencia CF \allowbreak \textless /{\color{ForestGreen}football\_team}\textgreater~in \textless {\color{RubineRed}location}\textgreater~Madrid \textless /{\color{RubineRed}location}\textgreater~\textless /response\textgreater}\end{tabular} \\ \midrule
\multicolumn{2}{c}{\textbf{Evaluation}} \\ \midrule
\textbf{Prompt}            & \begin{tabular}[c]{@{}l@{}}\parbox[t]{5in}{Your task is to perform Named Entity Recognition by wrapping each entity from the input text with label tags. The set of label tags you can use are: {\color{ForestGreen}football\_team}, and {\color{RubineRed}location}. Wrap all your response between \textless response\textgreater~... \textless/response\textgreater. Here are some examples:\\ \\ \underline{Text}: OpenAI is developing GPT-5.\\\underline{Labels}: \textless response\textgreater~OpenAI is developing GPT-5 \textless /response\textgreater\\ \underline{Text}: Real Madrid lose against Valencia CF in Madrid.\\\underline{Labels}: \textless response\textgreater~\textless {\color{ForestGreen}football\_team}\textgreater~Real Madrid \textless /{\color{ForestGreen}football\_team}\textgreater~lose against \textless {\color{ForestGreen}football\_team}\textgreater~Valencia CF \allowbreak \textless /{\color{ForestGreen}football\_team}\textgreater~in \textless {\color{RubineRed}location}\textgreater~Madrid \textless /{\color{RubineRed}location}\textgreater~\textless /response\textgreater \\ \underline{Text}: FC Barcelona is playing in Barcelona}\end{tabular} \\ \midrule
\textbf{LLM output}        & \begin{tabular}[c]{@{}l@{}}\parbox[t]{5in}{\textless response\textgreater~\textless {\color{ForestGreen}football\_team}\textgreater~FC Barcelona \textless /{\color{ForestGreen}football\_team}\textgreater~is playing in \textless {\color{RubineRed}location}\textgreater~Barcelona \textless /{\color{RubineRed}location}\textgreater~\textless response\textgreater}\end{tabular} \\ \midrule
\textbf{Parsed output}     & \begin{tabular}[c]{@{}l@{}}\parbox[t]{5in}{[{\color{ForestGreen}B-football\_team}, {\color{ForestGreen}I-football\_team}, O, O, O, {\color{RubineRed}B-location}]}\end{tabular} \\ \bottomrule
\end{tabular}%
\caption{\label{tab:ner-example} Example of dataset preparation and evaluation of a NER task using the proposed annotation schema. The prompt includes two in-context examples and the instance to predict is added below them. The underlines and colorings have been added for easier understanding but are not included as part of the inputs or outputs.}
\end{table}

\begin{table}[!t]
\centering
\resizebox{1.00\textwidth}{!}{%
\footnotesize
\begin{tabular}{lllllll}
\toprule
\textbf{Source} &   \textbf{Task} & \textbf{Subtask} & \textbf{Year} &  
\textbf{Language} & $|\mathcal{L}|$ & \textbf{Category} \\
\midrule
TweetLID & TweetLID \cite{zubiaga2016tweetlid} & Language dentification & 2014 & es & 7 & Language Identification \\
TASS & TASS \cite{garcia2020overview} & Emotion analysis & 2020 & es & 7 & Sentiment and Emotion Analysis \\
& TASS \cite{garcia2020overview} & Sentiment analysis & 2020 & es-\{UY, MX, ES, CR, PE\} & 3 & Sentiment and Emotion Analysis \\
PAN & Author Profiling \cite{rangel2015overview} & Age detection & 2015 & es & 4 & Author Profiling \\
& Author Profiling \cite{rangel2017overview} & Gender detection &  2017 & es & 2 & Author Profiling \\
IberEval & MultiStanceCat \cite{taule2018overview} & Stance detection & 2018 & \{es, ca\} & 3 & Stance Detection \\
IberLEF & HAHA \cite{chiruzzo2019overview} & Humor detection & 2019 & es & 2 & Humor Detection \\
& IroSvA \cite{ortega2019overview} & Irony detection & 2019 & es-\{CU, MX, ES\} & 2 & Irony and Sarcasm Detection \\
& MEX-A3T \cite{aragon2019overview} & Aggressiveness detection & 2019 & es-MX & 2 & Toxicity and Harmful Language Detection \\
& DETOXIS \cite{PLN6390} & Aggressiveness detection & 2021 & es & 2 & Toxicity and Harmful Language Detection \\
& DETOXIS \cite{PLN6390} & Improper language detection & 2021 & es & 2 & Toxicity and Harmful Language Detection  \\
& DETOXIS \cite{PLN6390} & Insult detection & 2021 & es & 2 & Toxicity and Harmful Language Detection  \\
& DETOXIS \cite{PLN6390} & Mockery detection & 2021 & es & 2 & Toxicity and Harmful Language Detection \\
& DETOXIS \cite{PLN6390} & Sarcasm detection & 2021 & es & 2 & Irony and Sarcasm Detection \\
& DETOXIS \cite{PLN6390} & Toxicity detection & 2021 & es & 2 & Toxicity and Harmful Language Detection \\
& EmoEvalEs \cite{PLN6385} & Emotion analysis & 2021 & es & 7 & Sentiment and Emotion Analysis \\
& EmoEvalEs \cite{PLN6385} & Offensiveness detection & 2021 & es & 2 & Toxicity and Harmful Language Detection \\
& EXIST \cite{PLN6389} & Sexism categorization & 2021 & es & 6 & Prejudice and Discrimination Detection \\
& EXIST \cite{PLN6389} & Sexism detection & 2021 & es & 2 & Prejudice and Discrimination Detection \\
& FakeDeS \cite{PLN6391} & Fake News detection & 2021 & es & 2 & Fake News Detection \\
& HAHA \cite{PLN6394} & Humor detection & 2021 & es & 2 & Humor Detection  \\
& MeOffendEs \cite{PLN6388} & Gender detection & 2021 & es & 2 & Author Profiling \\
& MeOffendEs \cite{PLN6388} & Offensiveness detection & 2021 & es & 4 & Toxicity and Harmful Language Detection \\
& Rest-Mex \cite{PLN6386} & Gender detection & 2021 & es-MX & 3 & Author Profiling \\
& Rest-Mex \cite{PLN6386} & Sentiment analysis & 2021 & es-MX & 5 & Sentiment and Emotion Analysis \\
& VaxxStance \cite{PLN6387} & Stance detection & 2021 & \{eu, es\} & 3 & Stance Detection  \\
& PAR-MEX \cite{PLN6445} & Paraphrase detection & 2022 & es & 2 & Paraphrase Detection \\
& Rest-Mex \cite{PLN6449} & Sentiment analysis & 2022 & es-MX & 5 & Sentiment and Emotion Analysis \\
& MentalRiskES \cite{PLN6564} & Depression categorization & 2023 & es & 4 & Mental Health Detection \\
& MentalRiskES \cite{PLN6564} & Depression detection & 2023 & es & 2 & Mental Health Detection \\
& MentalRiskES \cite{PLN6564} & Eating disorder detection & 2023 & es & 2 & Mental Health Detection \\
& HUHU \cite{PLN6568} & Fatphobia detection & 2023 & es & 2 & Prejudice and Discrimination Detection \\
& HUHU \cite{PLN6568} & Humor detection & 2023 & es & 2 & Humor Detection \\
& HUHU \cite{PLN6568} & LGBTIQ prejudice detection & 2023 & es & 2 & Prejudice and Discrimination Detection \\
& HUHU \cite{PLN6568} & Racial prejudice detection & 2023 & es & 2 & Prejudice and Discrimination Detection \\
& HUHU \cite{PLN6568} & Women prejudice detection & 2023 & es & 2 & Prejudice and Discrimination Detection \\
& DETESTS-Dis \cite{PLN6620} & Stereotype detection & 2024 & es & 2 & Prejudice and Discrimination Detection \\
& IberAuTexTification \cite{PLN6628} & MGT attribution & 2024 & \{en, es, eu, pt, ca, gl\} & 6 & MGT Detection and Attribution \\
& IberAuTexTification \cite{PLN6628} & MGT detection & 2024 &\{en, es, eu, pt, ca, gl\} & 2 & MGT Detection and Attribution \\
General & PAWS \cite{yang-etal-2019-paws,vladu2022proxecto,gonzalez-agirre-etal-2024-building-data} & Paraphrase Detection & 2019 & \{es, gl, pt, ca\} & 2 & Paraphrase detection \\
& XLSum \cite{hasan-etal-2021-xl} & Text summarization & 2021 & \{es, pt\} & - & Text Summarization \\
& Parafraseja \cite{proyecto_aina} & Paraphrase detection & 2022 & ca & 2 & Paraphrase Detection \\ 
& BEC \cite{urbizu2022basqueglue} & Sentiment analysis & 2024 & eu & 3 & Sentiment and Emotion Analysis \\
& BHTC \cite{urbizu2022basqueglue} & Topic classification & 2024 & eu & 12 & Topic Classification \\
& caBreu \cite{gonzalez-agirre-etal-2024-building-data} & Text summarization & 2024 & ca & - & Text Summarization \\
& ClinDiagnosES \cite{clindiagnoses24} & Topic classification & 2024 & es & 8 &  Topic Classification \\
& FMTODeu \cite{urbizu2022basqueglue} & Intent classification & 2024 & eu & 12 & Intent Classification \\
& HateCheck \cite{rottger-etal-2022-multilingual} & Hate speech detection & 2024 & pt & 2 & Toxicity and Harmful Lang Detection \\
\bottomrule
\end{tabular}
}
\caption{
 Industry-relevant tasks and datasets in IberBench. $|\mathcal{L}|$ refers to the number of labels or candidate answers if applicable. We indicate language varieties in brackets. If no variety is specified, it means the source does not provide information about which varieties are included in the dataset.
} \label{tab:datasets-description-industry-relevant-task}
\end{table}

\begin{table}[!t]
\centering
\resizebox{1.00\textwidth}{!}{%
\footnotesize
\begin{tabular}{lllllll}
\toprule
\textbf{Source} &   \textbf{Task} & \textbf{Subtask} & \textbf{Year} &  
\textbf{Language} & $|\mathcal{L}|$ & \textbf{Category} \\
\midrule
IberLEF & ADoBo \cite{PLN6396} & Lexical borrowing chunking & 2021 & es & 2 & Lexical Analysis \\
General & TE-ca \cite{armengol-estape-etal-2021-multilingual} & Textual entailment & 2021 & ca & 2 & Textual Entailment  \\
& OpenBookQA \cite{proyecto_ilenia, proyecto_aina} & Question answering & 2022 & \{es, ca\} & 4 & Question Answering \\
& ARC \cite{baucells-etal-2025-iberobench,proyecto_aina} & Question answering & 2024 & \{eu, ca\} & 4 & Question Answering  \\
& Belebele \cite{bandarkar-etal-2024-belebele} & Reading comprehension & 2024 & \{eu, es, pt, ca\} & 4 & Reading Comprehension  \\
& CoLA \cite{bel-etal-2024-escola,PLN6609,vladu2022proxecto}& Linguistic acceptability & 2024 & \{es, gl, ca\} & 2 & Linguistic Acceptability \\
& COPA \cite{baucells-etal-2025-iberobench,proyecto_ilenia,gonzalez-agirre-etal-2024-building-data}& Commonsense reasoning & 2024 & \{eu, es, ca\} & 2 & Commonsense reasoning \\
& EusExams \cite{etxaniz-etal-2024-latxa} & Question answering & 2024 & \{eu, es\} & 4 & Question Answering \\
& EusProficiency \cite{etxaniz-etal-2024-latxa} & Proficiency evaluation & 2024 & eu & 4 & Proficiency Evaluation  \\
& EusReading \cite{etxaniz-etal-2024-latxa}& Reading comprehension & 2024 & eu & - & Reading Comprehension \\
& EusTrivia \cite{etxaniz-etal-2024-latxa} & Question answering & 2024 & eu & 4 & Question Answering  \\
& EusTrivia \cite{etxaniz-etal-2024-latxa} & Topic classification & 2024 & eu & 5 & Topic Classification  \\ 
& PIQA \cite{baucells-etal-2025-iberobench} & Commonsense reasoning & 2024 & eu & 2 & Commonsense Reasoning \\
& QNLI \cite{urbizu2022basqueglue} & Textual entailment & 2024 & eu & 2 & Textual Entailment \\
& TELEIA \cite{spanish_benchmark_teleia} & Proficiency evaluation & 2024 & es & 4 & Proficiency Evaluation  \\
& XNLI \cite{conneau2018xnli,vladu2022proxecto,gonzalez-agirre-etal-2024-building-data} & Textual Entailment & 2024 & \{es, gl, ca\} & 2 & Textual entailment  \\
& XStoryCloze \cite{vladu2022proxecto, proyecto_aina} & Question answering & 2024 & \{gl, pt, ca\} & 2 & Question Answering  \\
\bottomrule
\end{tabular}
}
\caption{
\label{tab:datasets-description-fundamental-relevant-task}
 Fundamental tasks and datasets in IberBench. $|\mathcal{L}|$ refers to the number of labels or candidate answers if applicable. We indicate language varieties in brackets. If no variety is specified, it means the source does not provide information about which varieties are included in the dataset. 
} 
\end{table}

We put together a total of 101 datasets, which are detailed in Tables \ref{tab:datasets-description-industry-relevant-task} and \ref{tab:datasets-description-fundamental-relevant-task}. These tables include information on the originating workshop, dataset or task name, year, language, language variety (if applicable), number of labels (if applicable), and citation. URL information is available in the Appendix (Tables \ref{tab:appendix-description-industry-relevant-task} and \ref{tab:appendix-description-fundamental-relevant-task}). Additionally, the datasets are organized into categories and assessed for relevance to facilitate broader insights following evaluation. Categories represent broad semantic groupings of tasks; for example, \textit{Gender detection} and \textit{Age detection} are both subcategories of \textit{Author profiling}. Relevance, on the other hand, differentiates between fundamental and industry-relevant tasks. Fundamental tasks are those that an LLM must be able to perform in order to demonstrate proficiency in language use, reasoning, and factual knowledge. Industry-relevant tasks, conversely, are those with economic significance, such as detecting hate speech or machine-generated text for content moderation, and performing sentiment analysis or author profiling for customer insights. Most industry-relevant tasks are sourced from workshops, while the majority of fundamental tasks originate from established LLM benchmarks (designated in Tables \ref{tab:datasets-description-industry-relevant-task} and \ref{tab:datasets-description-fundamental-relevant-task} as ``General'' in the source column). This highlights two key points: (i) workshops are closely aligned with industry needs of NLP-focused companies, and (ii) existing LLM benchmarks tend to overlook industry-relevant tasks in favor of evaluating the fundamental capabilities of LLMs.

\begin{table}[t!]
\centering
\resizebox{\textwidth}{!}{%
\begin{tabular}{lrrrrrrrrrrrrrr}
\toprule
& $|\mathcal{D}|$ & $|\mathcal{S}|$ & eu & ca & gl & pt & en & \multicolumn{7}{c}{es} \\
\cmidrule(lr){9-15}
& & & & & & & & PE & UY & CR & AMB & CU & MX & ES \\
\midrule
MGT Detection and Attribution & 12 & 67.3 & 8.6 & 12.6 & 2.6 & 13.7 & 15.7 & 0 & 0 & 0 & 13.8 & 0 & 0 & 0 \\
Author Profiling & 4 & 54.4 & 0 & 0 & 0 & 0 & 0 & 0 & 0 & 0 & 52.2 & 0 & 2.2 & 0 \\
Question Answering & 10 & 43.6 & 18.7 & 3.2 & 1.5 & 1.5 & 0 & 0 & 0 & 0 & 18.7 & 0 & 0 & 0 \\
Toxicity and Harmful Language Detection & 9 & 23.9 & 0 & 0 & 0 & 3.7 & 0 & 0 & 0 & 0 & 19.7 & 0 & 0.6 & 0 \\
Language Identification & 1 & 18.4 & 0 & 0 & 0 & 0 & 0 & 0 & 0 & 0 & 18.4 & 0 & 0 & 0 \\
Textual Entailment & 5 & 14.9 & 0.2 & 7.1 & 5.0 & 0 & 0 & 0 & 0 & 0 & 2.5 & 0 & 0 & 0 \\
Humor Detection & 3 & 12.8 & 0 & 0 & 0 & 0 & 0 & 0 & 0 & 0 & 12.8 & 0 & 0 & 0 \\
Text Summarization & 3 & 12.2 & 0 & 0.3 & 0 & 7.2 & 0 & 0 & 0 & 0 & 4.7 & 0 & 0 & 0 \\
Paraphrase Detection & 6 & 12.1 & 0 & 6.0 & 2.0 & 2.0 & 0 & 0 & 0 & 0 & 2.0 & 0 & 0.1 & 0 \\
Linguistic Acceptability & 3 & 11.9 & 0 & 9.2 & 1.7 & 0 & 0 & 0 & 0 & 0 & 1.1 & 0 & 0 & 0 \\
Prejudice and Discrimination Detection & 7 & 9.6 & 0 & 0 & 0 & 0 & 0 & 0 & 0 & 0 & 9.6 & 0 & 0 & 0 \\
Proficiency Evaluation & 2 & 5.3 & 5.2 & 0 & 0 & 0 & 0 & 0 & 0 & 0 & 0.1 & 0 & 0 & 0 \\
Reading Comprehension & 5 & 3.9 & 1.2 & 0.9 & 0 & 0.9 & 0 & 0 & 0 & 0 & 0.9 & 0 & 0 & 0 \\
Topic Classification & 3 & 3.6 & 3.6 & 0 & 0 & 0 & 0 & 0 & 0 & 0 & 0.06 & 0 & 0 & 0 \\
Commonsense Reasoning & 4 & 3.3 & 2.3 & 0.5 & 0 & 0 & 0 & 0 & 0 & 0 & 0.5 & 0 & 0 & 0 \\
Stance Detection & 4 & 3.3 & 0.3 & 1.2 & 0 & 0 & 0 & 0 & 0 & 0 & 1.8 & 0 & 0 & 0 \\
Irony and Sarcasm Detection & 4 & 2.7 & 0 & 0 & 0 & 0 & 0 & 0 & 0 & 0 & 0.9 & 0.6 & 0.6 & 0.6 \\
Sentiment and Emotion Analysis & 10 & 26.4 & 1.3 & 0 & 0 & 0 & 0 & 1.4 & 1.4 & 1.4 & 2.5 & 0 & 16.7 & 1.7 \\
Lexical Analysis & 1 & 1.8 & 0 & 0 & 0 & 0 & 0 & 0 & 0 & 0 & 1.8 & 0 & 0 & 0 \\
Intent Classification & 1 & 1.1 & 1.1 & 0 & 0 & 0 & 0 & 0 & 0 & 0 & 0 & 0 & 0 & 0 \\
Fake News Detection & 1 & 0.6 & 0 & 0 & 0 & 0 & 0 & 0 & 0 & 0 & 0.66 & 0 & 0 & 0 \\
Mental Health Detection & 3 & 0.4 & 0 & 0 & 0 & 0 & 0 & 0 & 0 & 0 & 0.4 & 0 & 0 & 0 \\
\midrule
\midrule
Industry NLP & 70 & 247.1 & 13.2 & 20.1 & 4.6 & 26.6 & 15.8 & 1.4 & 1.4 & 1.2 & 139.6 & 0.6 & 20.1 & 2.3 \\
Fundamental NLP & 31 & 86.4 & 29.4 & 20.9 & 8.2 & 2.4 & 0 & 0 & 0 & 0 & 25.6 & 0 & 0 & 0 \\
\midrule
\midrule
Total & 101 & 333.5 & 42.6 & 41.0 & 12.8 & 29.0 & 15.8 & 1.4 & 1.4 &  1.2 & 165.2 & 0.6 & 20.1 & 2.3 \\ 
\bottomrule
\end{tabular}%
}
\caption{\label{tab:datasets-statistics}Statistics of IberBench. $|\mathcal{D}|$ and $|\mathcal{S}|$ refer to number of datasets and samples respectively. The remaining columns show the number of samples per language in thousands. ``AMB'' stands for ``ambiguous'' and refers to potential mixes of language variations that are not explicitly identified by dataset creators.}
\end{table}

To complement the dataset descriptions, Table \ref{tab:datasets-statistics} presents total and aggregated statistics per category and relevance. IberBench comprises 101 datasets, distributed across 22 categories, with over 333 thousand samples, which are widely dispersed across 12 Iberian languages.\footnote{We include English for two reasons. First, to leverage the efforts of the Iberian NLP community in creating English datasets. Second, because English is spoken in Gibraltar, a region geographically connected to the Iberian Peninsula. Only English datasets created within Iberian evaluation campaigns are included.} Ninety-seven of these datasets are framed as text classification, 3 of them as text generation, and 1 as sequence labeling.

Regarding languages, Spanish is the most represented, with 192.3 thousand samples, accounting for approximately 60\% of the benchmark, followed by Basque and Catalan, which together represent around 12\%. Galician is the least represented, with only 4\% of the total. 
Mexican (6.0\%) and Spanish from Spain (0.6\%) are the most represented among the Spanish varieties, without considering the ``ambiguous'' variety (49.5\%) that potentially contain mixed variations. Other varieties including Peruvian, Uruguayan, Costa Rican, and Cuban, require more effort to generate meaningful resources for the NLP scientific community and achieve better representation. Still, most Spanish varieties are primarily included in a few task categories, such as Sentiment and Emotion Analysis, and Irony and Sarcasm Detection.

Concerning task categories, the most populated ones are Sentiment and Emotion Analysis (7.9\%), Machine-Generated Text Detection and Attribution (20.2\%), Author Profiling (16.3\%), and Question Answering (13.1\%). The first three categories are mainly derived from the TASS, IberLEF, and PAN workshops, which have historically focused on addressing these tasks across a wide range of Iberian languages. In contrast, Question Answering is almost entirely composed of datasets from LLM benchmarks. The two categories with the most samples are Machine-Generated Text (MGT) Detection and Attribution, and Author Profiling. The datasets used in these tasks are typically generated automatically, either by using LLMs to generate text or inferring user demographics from social media. Mental Health and Fake News Detection stand out among the less populated task categories. Despite being critically important for industry NLP, these categories are inherently complex, making it challenging to create datasets to study them effectively. There is a high imbalance in terms of (task category, language) pairs. For instance, Mental Health Detection only comprises data in Spanish but Sentiment and Emotion Analysis comprises data in near all Iberian language. This implies that LLMs might appear weaker or stronger in cross-lingual and cross-task comparisons depending on how well they perform on overrepresented or underrepresented combinations. We expect that in the future, novel and comparable datasets across languages may arise, that can incrementally added into IberBench.

IberBench includes 70 datasets for industry-relevant tasks and 31 for fundamental ones. Of the total, 247.1 thousand samples are from industry-relevant tasks, while 86.4 thousand are from fundamental tasks, representing 74\% and 26\% of the total, respectively. It has been proven that recent LLMs demonstrate strong capabilities in natural language understanding, generation, and knowledge. However, their use in industry-relevant tasks that matter most to NLP-focused companies is still largely unexplored. Therefore, we consider these proportions to represent a balanced trade-off between evaluating the language and knowledge capabilities of LLMs and assessing their practical usefulness in real-world scenarios.

There are several areas for improvement where the Iberian NLP community should focus future efforts on dataset creation. The task-language table is sparse, emphasizing the need to develop more datasets. Industry-relevant tasks and certain Spanish varieties, particularly Peruvian, Uruguayan, and Costa Rican Spanish, show the most gaps. Additionally, we observe the absence of specific language varieties, such as Brazilian Portuguese and Argentinian Spanish. Ninety-seven of the datasets included in IberBench are framed as text classification tasks, 3 as text generation (which only include text summarization), and 1 as sequence labeling. These proportions reflect the greater difficulty involved in creating text generation and sequence labeling datasets, as well as the complexities associated with their evaluation. As a result, there is a clear need for more efforts to develop datasets for more diverse task types.

In this section, we have mainly discussed the retrospective collection of datasets. However, IberBench also facilitates prospective dataset collection. When the organization approves new datasets, a designated team member runs the workflow outlined in this section. This process includes downloading the dataset, specifying a configuration file to guide the normalization process,\footnote{We provide examples of configurations to normalize datasets at \url{https://github.com/IberBench/dataset-preparation/tree/main/configs}} and uploading the processed dataset to the private HuggingFace repository.

\subsection{LLM Evaluation}

IberBench is designed to exclusively evaluate autoregressive LLMs. This choice is motivated by their substantial flexibility in addressing a wide range of NLP tasks and their significant traction in both academic research and industrial applications. The evaluation of an autoregressive LLM on a task $\mathcal{T}$ proceeds as follows. Let $f: \mathcal{P} \times \mathcal{X} \rightarrow \tilde{\mathcal{P}}$ denote a verbalization function that formats an input $x \in \mathcal{X}$ into a prompt template, where $p \in \mathcal{P}$ is a prompt template containing a task description—potentially including few-shot exemplars. Let $\mathcal{D} = \{(x_1, y_1), \ldots, (x_n, y_n) : x_i \in \mathcal{X},\ y_i \in \mathcal{Y}\}$ denote a test dataset comprising input-output pairs, then a set of instantiated prompts is constructed as $\mathcal{S} = \{s_1, \ldots, s_n : s_i = f(p_i, x_i) \in \tilde{\mathcal{P}}\}$. The LLM is then prompted with each $s_i \in \mathcal{S}$ and generates a corresponding output that is subsequently compared with the reference output $y_i$. The nature of the generated output depends on the type of task; in this work, we restrict our focus to text classification, text generation, and sequence labeling. For classification tasks the label set is predefined, allowing the model to compute the likelihood of each candidate label. The label with the highest likelihood is selected, as formalized in Equation~\ref{eq:llm_classification},

\begin{equation}
\label{eq:llm_classification}
\hat{y} = \underset{y\in\mathcal{Y}}{\textrm{argmax}} \prod_{j=0}^{|y|} p_{LLM}(y_j \mid s_i, y_{<j})
\end{equation}

\noindent where $p_{\text{LLM}}(y_j \mid s_i, y_{<j})$ denotes the probability assigned by the LLM to the $j$-th token of the label, conditioned on the instantiated prompt $s_i$ and the preceding tokens of the label. For generation tasks, we employ greedy decoding to ensure reproducibility. Under this decoding strategy, the LLM generates $l$ tokens in an output sequence $\hat{y} = (\hat{y}_1, \ldots, \hat{y}_l)$, where each token $\hat{y}_j$ is produced according to Equation~\ref{eq:llm_generation},

\begin{equation}
\label{eq:llm_generation}
 \hat{y}_j = \underset{w\in\mathcal{V}}{argmax}\;p_{LLM}(w \mid s_i, \hat{y}_{<j})
\end{equation}

\noindent where $\mathcal{V}$ denotes the vocabulary of the LLM. This formulation is employed in our evaluation pipeline for both open-ended text generation tasks such as summarization, and for sequence labeling tasks that follow the annotation scheme outlined in Section~\ref{sec:datasets}. In open-ended text generation, the LLM continues generating tokens until an end-of-sequence token is produced, whereas in sequence labeling tasks, generation halts upon emitting the \texttt{<response>} tag.

Once the outputs for all input texts are generated by the LLM, we apply standard evaluation metrics to assess model performance. For classification tasks, we report the Macro-F$_1$ score to mitigate the impact of label imbalance in the test set. For generation tasks, we adopt ROUGE-1 \cite{lin-2004-rouge}, which measures unigram overlapping between the generated outputs and the reference texts. In the absence of artifact-free, perfectly human-aligned, and efficient automatic evaluation metrics for text summarization \cite{he-etal-2023-blind}, we opt for ROUGE-1 due to its widespread use in prior LLM benchmarks for evaluating text generation capabilities \cite{laleaderboard2024}. For sequence labeling tasks, we report the F$_1$ score as computed by the \texttt{seqeval} library \cite{seqeval}, a widely adopted tool for evaluating model performance on chunking tasks.

All evaluations in IberBench are conducted using \texttt{lm-eval-harness}, a well-established, extensively tested, open-source framework for LLM evaluation. This framework has served as the evaluation backbone for numerous leaderboards, becoming the reference framework in the scientific community due to its simplicity and flexibility. It facilitates the evaluation of LLMs from various providers with minimal effort, requiring only a YAML configuration file to specify the dataset, prompt, and generation hyperparameters,\footnote{We release all the prompts we use to evaluate LLMs at \url{https://github.com/IberBench/iberbench-evaluation/tree/main/lm_eval/tasks/iberbench}} supporting a wide range of setups and hardware configurations. For IberBench, we adapted \texttt{lm-eval-harness} for three primary purposes: managing sequence-labeling tasks, enabling incremental evaluation through caching, and supporting on-premise execution.

To evaluate sequence-labeling tasks, (i) we designed a custom annotation schema, (ii) implemented label reconstruction from generations, (iii) integrated the \texttt{seqeval} metric into the framework, and (iv) prepared YAML configuration files and guidelines for its application.\footnote{We show an example of YAML for a sequence-labeling task at \url{https://github.com/IberBench/iberbench-evaluation/blob/main/lm_eval/tasks/iberbench/iberlef-adobo-lexical_borrowing_chunking-2021-spanish.yaml}} While we store all LLM evaluation requests and results in HuggingFace repositories, \texttt{lm-eval-harness} does not natively interact with our results datasets and re-computes the results each time a task is executed with an LLM. This is not desirable for incremental evaluation, and we address it by developing custom evaluation code within \texttt{lm-eval-harness} that (i) identifies models pending evaluation, including both new models and existing ones that have not been evaluated with recently added tasks, (ii) checks which tasks have already been computed by a model, (iii) runs the model only on tasks not yet evaluated, and (iv) caches the results in HuggingFace repositories.\footnote{The evaluation script for incremental evaluation can be accessed at \url{https://github.com/IberBench/iberbench-evaluation/blob/main/scripts/iberbench/eval_iberbench.py}} Finally, most of the leaderboards hosted on HuggingFace rely on the \texttt{lm-eval-harness} pipeline integrated into HuggingFace's servers, which are based on CPU. This setup is insufficient for evaluating LLMs with large parameter sizes at scale. Instead, IberBench performs evaluations on-premise, on a custom  server equipped with two 48GB A6000 Nvidia GPUs, evaluating unquantized models up to 14 billion parameters in approximately 12 hours per model.

The organization is responsible for running the custom evaluation code on-premise whenever a batch of models or datasets is approved. For a new model, the evaluation script is executed, automatically assessing the model across all existing datasets. For a new dataset, the organization creates a YAML configuration file and runs the evaluation module, which updates the results of all existing models with the new dataset.

\section{Evaluation} \label{sec:evaluation}
We conducted an extensive evaluation of multilingual LLMs, including those specialized on some Iberian languages, on the IberBench benchmark. This section details the models assessed so far, presents the evaluation results, and discusses the key insights derived from the analysis.

\subsection{Models}

The ecosystem of LLMs—particularly high-quality ones—for Iberian languages remains limited. The majority of existing models rely on further pre-training of multilingual LLMs to improve language adaptation, or on enhancing alignment with human preferences in the target languages through supervised fine-tuning and reinforcement learning techniques \cite{NEURIPS2023_a85b405e, schulman2017proximalpolicyoptimizationalgorithms}. Only a small number of models have been pre-trained from scratch with Iberian languages on mind, most notably Salamandra \cite{gonzalezagirre2025salamandratechnicalreport} and EuroLLM \cite{MARTINS202553}.

In IberBench we aim to evaluate a broad spectrum of LLMs, which include (i) multilingual LLMs not specialized in Iberian languages that are close to the state of the art in English, (ii) LLMs pre-trained from scratch on most Iberian languages, and (iii) LLMs obtained by adapting existing multilingual models to one or more Iberian languages. For the first group, we include Microsoft's \texttt{phi-4} \cite{abdin2024phi4technicalreport} and \texttt{phi-4-mini-instruct} \cite{microsoft2025phi4minitechnicalreportcompact}, Meta's Llama 3.1 and 3.2 families \cite{grattafiori2024llama3herdmodels}, Alibaba's Qwen 2.5 family \cite{qwen2025qwen25technicalreport}, and \texttt{Mistral-7B-Instruct-v0.3} \cite{jiang2023mistral7b}.\footnote{We also included \texttt{gpt2} \cite{radford2019language} and \texttt{gemma-2-2b-it} \cite{gemmateam2024gemma2improvingopen} as functional baselines. They were trained exclusively on English and thus serves as low-bar baselines for assessing the relative performance of more recent models on other Iberian languages.} In the second group, we include Salamandra \cite{gonzalezagirre2025salamandratechnicalreport} and EuroLLM \cite{MARTINS202553} families. For the third group, we include \texttt{RigoChat-7b-v2} for Spanish \cite{gómez2025rigochat2adaptedlanguage}, \texttt{Latxa-Llama-3.1-8B-Instruct} for Basque \cite{etxaniz-etal-2024-latxa}, \texttt{CataLlama-v0.2-Instruct-SFT} for Catalan \cite{catallama}, \texttt{sabia-7b} for Portuguese \cite{10.1007/978-3-031-45392-2_15}, and \texttt{Aitana-6.3B} for Catalan/Valencian \cite{aitana}. The selected collection encompasses both base models pre-trained solely for causal language modeling and instruction-tuned models designed for instruction following and conversational tasks. Model sizes range from 100 million to 14 billion parameters, covering scenarios with low to moderate computational requirements for deployment and inference.

\begin{table}[t]
\centering
\resizebox{\textwidth}{!}{%
\begin{tabular}{@{}llllll@{}}
\toprule
Model Name & Type & Num. Params (billions) & Pre-training Languages & Fine-tuning Languages\\
\midrule
\texttt{phi-4} \cite{abdin2024phi4technicalreport} & \includegraphics[height=1em]{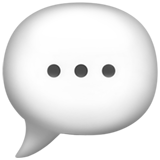} & 14.1 & es, en, pt & \includegraphics[height=1em]{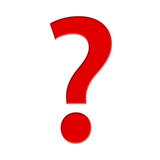} \\
\texttt{EuroLLM-9B-Instruct} \cite{MARTINS202553} & \includegraphics[height=1em]{emojis/chat.png} & 9.1 & es, en, ca, pt, gl& es, en, ca, pt, gl\\
\texttt{EuroLLM-9B} \cite{MARTINS202553} & \includegraphics[height=1em]{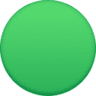} & 9.1 & es, en, ca, pt, gl& - \\
\texttt{CataLlama-v0.2-Instruct-SFT} \cite{catallama} & \includegraphics[height=1em]{emojis/chat.png} & 8.0 & es, en, pt & ca \\
\texttt{Llama-3.1-8B-Instruct} \cite{grattafiori2024llama3herdmodels} & \includegraphics[height=1em]{emojis/chat.png} & 7.5 & es, en, pt & es, en, pt \\
\texttt{Latxa-Llama-3.1-8B-Instruct} \cite{etxaniz-etal-2024-latxa}  & \includegraphics[height=1em]{emojis/chat.png} & 7.5 & es, en, pt & eu \\
\texttt{Qwen2.5-7B-Instruct} \cite{qwen2025qwen25technicalreport} & \includegraphics[height=1em]{emojis/chat.png} & 7.1 & es, en, pt & \includegraphics[height=1em]{emojis/question.png} \\
\texttt{RigoChat-7b-v2} \cite{gómez2025rigochat2adaptedlanguage} & \includegraphics[height=1em]{emojis/chat.png} & 7.1 & es, en, pt & es \\
\texttt{Mistral-7B-Instruct-v0.3} \cite{jiang2023mistral7b} & \includegraphics[height=1em]{emojis/chat.png} & 7.1 & es, en & \includegraphics[height=1em]{emojis/question.png} \\
\texttt{salamandra-7b-instruct} \cite{gonzalezagirre2025salamandratechnicalreport} & \includegraphics[height=1em]{emojis/chat.png} & 6.7 & es, en, ca, pt, gl, eu & es, en, ca, pt, gl, eu \\
\texttt{sabia-7b} \cite{10.1007/978-3-031-45392-2_15}  & \includegraphics[height=1em]{emojis/base.png} & 6.7 & es, en, pt & - \\
\texttt{Aitana-6.3B} \cite{aitana} &  \includegraphics[height=1em]{emojis/base.png} & 6.3 & es, en, ca & - \\
\texttt{Qwen2.5-3B-Instruct} \cite{qwen2025qwen25technicalreport}  & \includegraphics[height=1em]{emojis/chat.png} & 3.1 & es, en, pt & \includegraphics[height=1em]{emojis/question.png} \\
\texttt{Llama-3.2-3B-Instruct} \cite{grattafiori2024llama3herdmodels}  & \includegraphics[height=1em]{emojis/chat.png} & 3.2 & es, en, pt & es, en, pt \\
\texttt{phi-4-mini-instruct} \cite{microsoft2025phi4minitechnicalreportcompact} & \includegraphics[height=1em]{emojis/chat.png} & 3.8 & es, en, pt & \includegraphics[height=1em]{emojis/question.png} \\
\texttt{gemma-2-2b-it} \cite{gemmateam2024gemma2improvingopen}  & \includegraphics[height=1em]{emojis/chat.png} & 2.6 & en & en \\
\texttt{salamandra-2b-instruct} \cite{gonzalezagirre2025salamandratechnicalreport}  & \includegraphics[height=1em]{emojis/chat.png} & 1.7 & es, en, ca, pt, gl, eu & es, en, ca, pt, gl, eu \\
\texttt{EuroLLM-1.7B-Instruct} \cite{MARTINS202553}  & \includegraphics[height=1em]{emojis/chat.png} & 1.7 & es, en, ca, pt, gl& es, en, ca, pt, gl \\
\texttt{EuroLLM-1.7B} \cite{MARTINS202553}  & \includegraphics[height=1em]{emojis/base.png} & 1.7 & es, en, ca, pt, gl& - \\
\texttt{Qwen2.5-1.5B-Instruct} \cite{qwen2025qwen25technicalreport}  & \includegraphics[height=1em]{emojis/chat.png} & 1.5 & es, en, pt & \includegraphics[height=1em]{emojis/question.png} \\
\texttt{Llama-3.2-1B-Instruct} \cite{grattafiori2024llama3herdmodels}  & \includegraphics[height=1em]{emojis/chat.png} & 1.2 & es, en, pt & es, en, pt \\
\texttt{Qwen2.5-0.5B-Instruct} \cite{qwen2025qwen25technicalreport}  & \includegraphics[height=1em]{emojis/chat.png} & 0.5 & es, en, pt & \includegraphics[height=1em]{emojis/question.png} \\
\texttt{gpt2} \cite{radford2019language}  & \includegraphics[height=1em]{emojis/base.png} & 0.1 & en & - \\
\bottomrule
\end{tabular}%
}
\caption{\label{tab:llms}LLMs evaluated in IberBench. Includes the model name, type, number of parameters in billions, and Iberian languages included in the pre-training and fine-tuning mixtures. \includegraphics[height=1em]{emojis/base.png} indicates base models solely pre-trained via causal language modeling; \includegraphics[height=1em]{emojis/chat.png} marks models further fine-tuned for instruction-following or chat tasks; \includegraphics[height=1em]{emojis/question.png} denotes cases where training details are unspecified in the source. Fine-tuning languages for base models are marked as ``-'', since no fine-tuning has been performed on these models. All LLMs can be accessed in the \href{https://huggingface.co/models}{HuggingFace Hub} searching by model name.}
\end{table}

Table \ref{tab:llms} provides an overview of the LLMs evaluated within IberBench. Of the models evaluated, 78.3\% are chat-based, while 21.7\% are base models. In terms of parameter size, 47.8\% of the models fall within the 0.1-5 billion parameter range, 47.8\% within 5.1-10 billion parameters, and only 4.4\% within 10.1-15 billion parameters. All the LLMs, except for \texttt{gpt2} and \texttt{gemma-2-2b-it}, support at least Spanish and English, with 82.6\% also covering Portuguese. The most widely covered co-official language in Spain is Catalan, included in either pre-training or fine-tuning by 34.8\% of the models, followed by Galician (26.1\%), and Basque (13.0\%). None of the LLMs specifically target Spanish varieties from Ibero-America, or this information is not reported in the source publication.

We evaluate these LLMs in a zero-shot setting, i.e., without providing any in-context examples. While few-shot learning can enhance performance on certain tasks, the effects of factors such as the selection, quality, ordering, format, and quantity of examples remain subjects of ongoing debate \cite{dong-etal-2024-survey}. Furthermore, few-shot prompting is not yet a widespread practice in industry, partly since many non-technical users are unfamiliar with this prompting technique and because obtaining realistic examples can be challenging for tasks such as depression detection. To maintain consistency across tasks, better reflect real-world scenarios, and assess the capabilities acquired during pre-training and fine-tuning without relying on pattern matching from provided examples, we opted to evaluate all LLMs in a zero-shot setting, with the only exception being sequence-labeling tasks. For these tasks, a few examples are necessary to guide the model in properly formatting the output sequences, and we used three shots for this purpose.

To provide a more grounded evaluation of the results obtained with LLMs, we compare their performance against random baselines. For classification tasks, the random baseline assigns a label randomly, following a uniform distribution. In generation tasks, the random baseline's prediction consists of two randomly selected sentences from the context (e.g., a document in text summarization) joined by a whitespace. For sequence-labeling tasks, the random baseline's prediction for an instance is generated by shuffling the reference sequence of labels.

As part of the IberBench release, we evaluated a selected set of LLMs consisting of the most prominent models for Iberian languages. Users also have the option to request the evaluation of additional LLMs for through the leaderboard interface.

\subsection{Results}

We analyze LLM performance across multiple dimensions, including overall performance and performance per task category, relevance, languages, and model types.

\begin{figure}[t]
\centering
\includegraphics[scale=0.473]{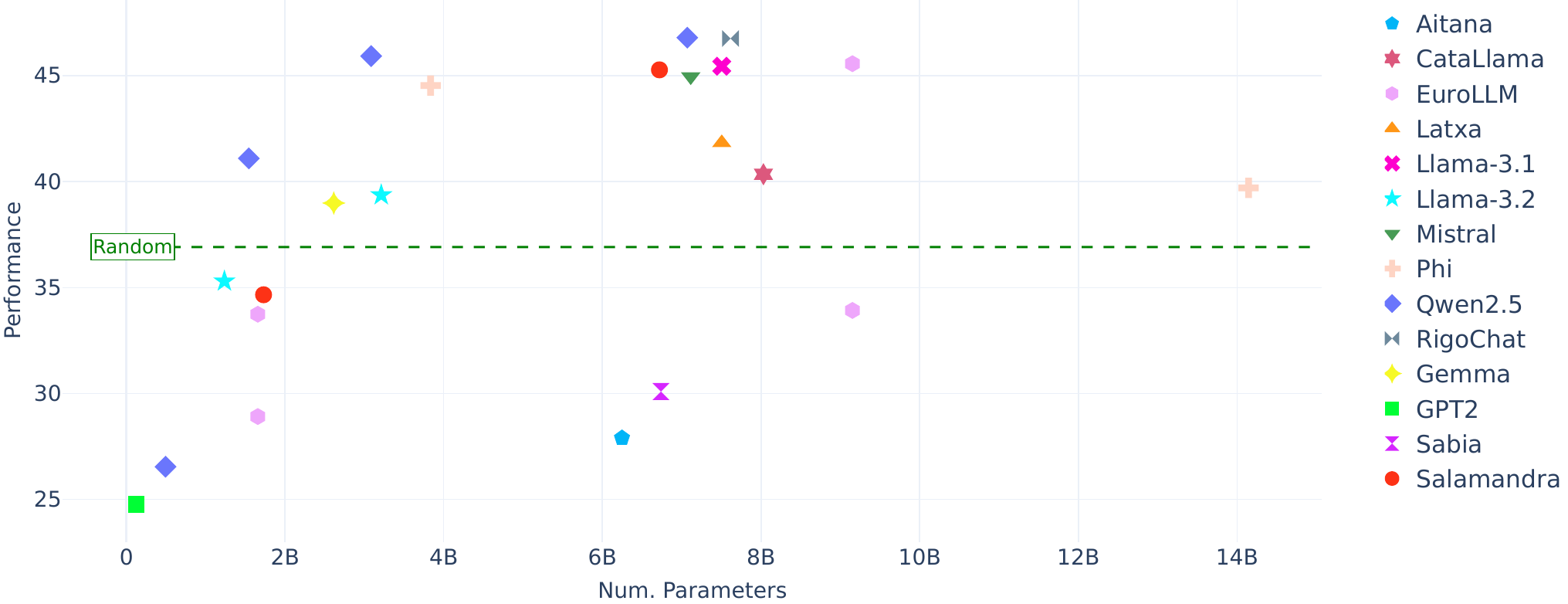}
\caption{\label{fig:all-performance-per-size} Performance per model size (number of parameters), averaged across all the languages and tasks. Legend shows model families, e.g., the Llama-3.2 family includes \texttt{Llama-3.2-1B-Instruct} and \texttt{Llama-3.2-3b-Instruct}, which are plotted with the same marker and color. The random baseline is shown as horizontal line.}
\end{figure}

\begin{figure}[t]
\centering
\includegraphics[scale=0.45]{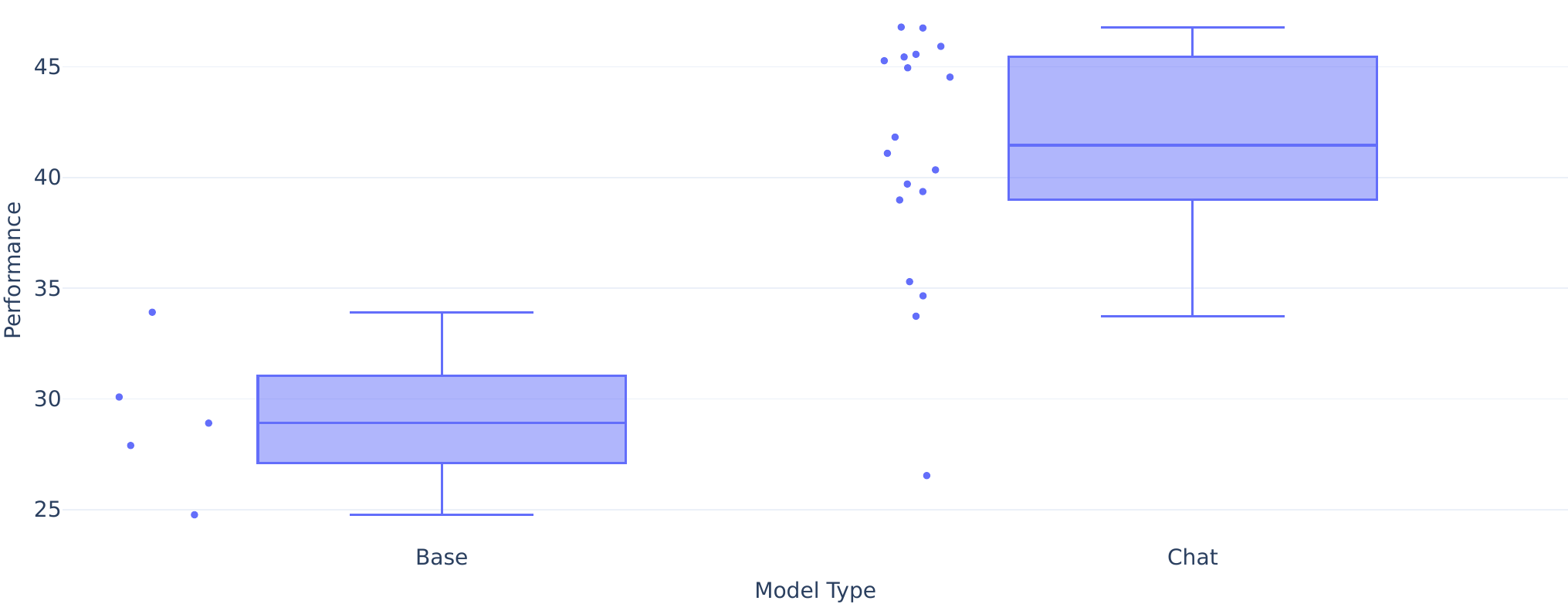}
\caption{\label{fig:all-performance-per-type} Averaged performance per model type.}
\end{figure}

\subsubsection{Overall Model Performance}
\label{sec:overall_performance}
We aim to identify which models are most competitive overall across tasks and languages in IberBench. To do this, we present the averaged performance across all datasets and languages together with model sizes in Figure \ref{fig:all-performance-per-size}.

\paragraph{The top-3 LLMs are all from the Qwen-2.5 family} \texttt{Qwen-2.5-7b-Instruct} dominates the benchmark with a mean score of 46.8\%, followed closely by \texttt{RigoChat-7b-v2} with 46.7\%, and \texttt{Qwen-2.5-3b-Instruct} with 45.9\%. These models are among the more recent in the ecosystem, with \texttt{RigoChat-7b-v2} being a fine-tuned version of \texttt{Qwen-2.5-7b-Instruct} optimized for Spanish. The Qwen-2.5 family is also one of the top-performing models in other multilingual benchmarks \cite{odesia2024, open-llm-leaderboard-v2}. 

\paragraph{Fine-tuning for language specialization can harm performance in other Iberian languages} We observe this behavior through \texttt{RigoChat-7b-v2} reporting better results than the original \texttt{Qwen-2.5-7b-Instruct} model. However, \texttt{RigoChat-7b-v2} performs slightly worse overall than \texttt{Qwen-2.5-7b-Instruct}. A similar observation can be done by comparing \texttt{Latxa-Llama-3.1-8B-Instruct}, an LLM optimized for Basque, with its base model \texttt{Llama-3.1-8b-Instruct}, which suggests that further specialization in a single language harms the performance of the model on other Iberian languages. 

\paragraph{Models from 3.1 to 10 billion parameters dominate the leaderboard} The best-performing models are primarily concentrated in the 3.1–10B parameter range. Models with fewer than 3B parameters, as well as \texttt{phi-4} (14B), often underperform relative to others. Based on scaling laws \cite{hoffmann2022training}, we expect that larger models will eventually outperform the current LLMs in the benchmark. In fact, scaling laws hold within families, e.g., \texttt{salamandra-7b-Instruct} performs better overall than \texttt{salamandra-2b-Instruct}. The same behavior is observed in Llama-3.2, Qwen-2.5, and EuroLLM families.

\paragraph{European LLMs fall short compared to popular multilingual LLMs} On average, European flagship models pre-trained from scratch with a focus on Iberian languages—such as \texttt{salamandra-7B-Instruct} and \texttt{EuroLLM-9B-Instruct}— perform on par with, or slightly below, models like \texttt{Qwen-2.5-7B-Instruct}, \texttt{Qwen-2.5-3B-Instruct}, and \texttt{Llama-3.1-8B-Instruct}, despite having comparable or larger parameter counts. Similarly, the smaller variants—\texttt{salamandra-2B-Instruct} and \texttt{EuroLLM-1.7B-Instruct}—fail to outperform \texttt{Llama-3.2-1B-Instruct}, even though they are slightly larger. Notably, none of these small models achieve better results than \texttt{gemma-2-2b-it}, despite it being an English monolingual model.

\paragraph{LLMs focused in Iberian languages are competitive only if they are instruction-tuned} Among the models that have been further pre-trained or fine-tuned from existing LLMs with a focus on specific Iberian languages—such as \texttt{Aitana-6.3B}, \texttt{sabia-7B}, \texttt{CataLlama-v0.2-Instruct-SFT}, \texttt{Latxa-Llama-3.1-8B-Instruct}, and \texttt{RigoChat-7B-v2}—only the latter three demonstrate competitive performance relative to the best-performing models. As evidenced in  Figure \ref{fig:all-performance-per-type}, this disparity may be attributed to the fact that \texttt{Aitana-6.3B} and \texttt{sabia-7B} are base models, which generally underperform compared to instruction-tuned models.

\paragraph{It is hard to beat the random baseline} Thirty-nine percent of the models performed below the random baseline. Among these, we identified all the base models and several instruction-tuned models with fewer than 2B parameters, including \texttt{Llama-3.2-1B-instruct}, \texttt{salamandra-2B-instruct}, \texttt{Qwen2.5-0.5B-instruct}, and \texttt{EuroLLM-1.7B-instruct}. This emphasizes the significance of model scale for instruction-tuned models: only those exceeding 2B parameters demonstrate competitive performance. Base models, even those surpassing the 6B parameter threshold, are not competitive. The difference between the best-performing LLM and the random baseline is approximately 10\%, highlighting that IberBench is still far from being adequately solved.

\subsubsection{Performance per Task}
\begin{figure}[t]
\centering
\includegraphics[scale=0.39]{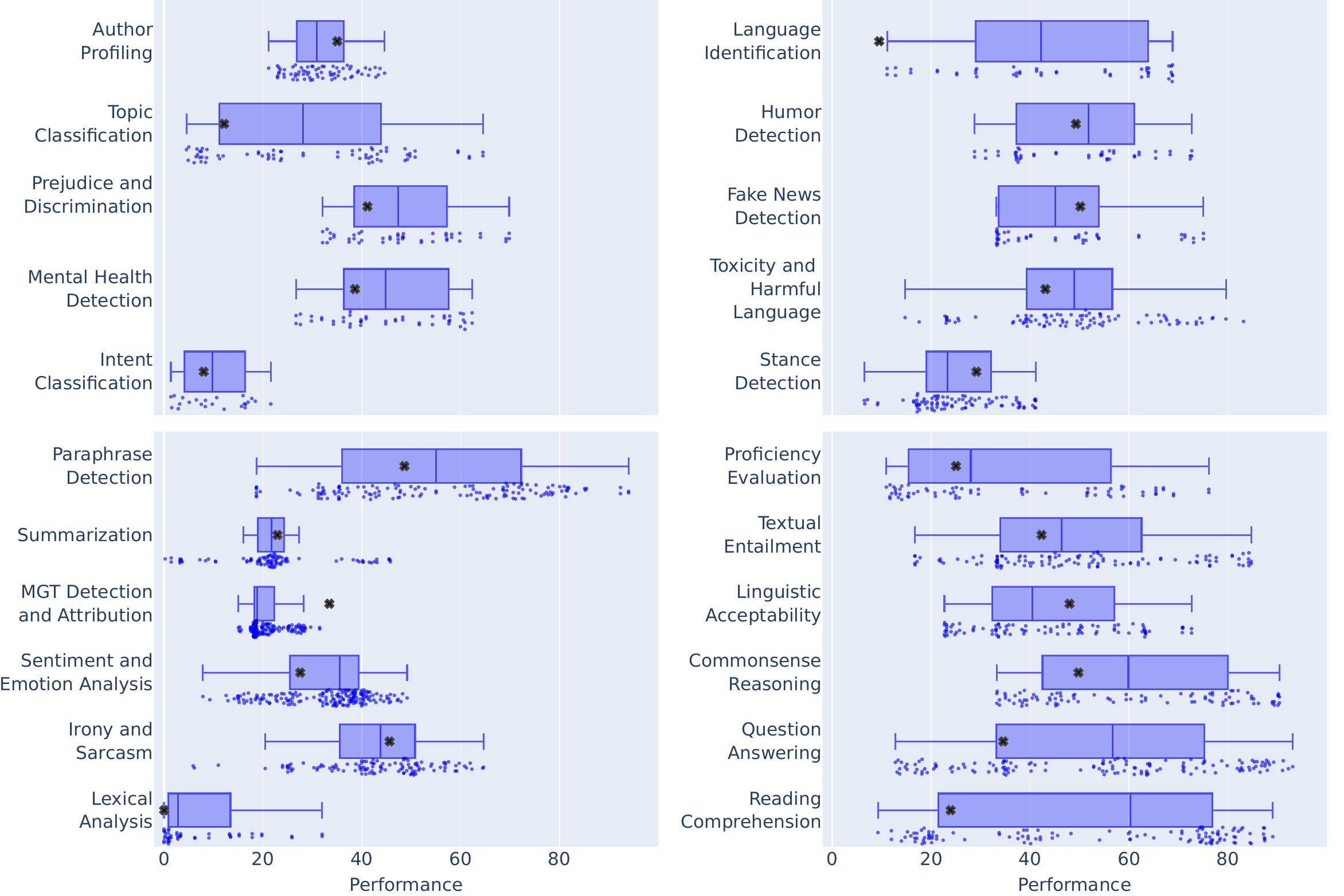}
\caption{\label{fig:all-performance-per-category} Performance per task category across all models and languages. The random baseline for each category is marked with a black cross.}
\end{figure}

We analyze performance by task to gain a better understanding of how well the LLMs handle each of them, specifically identifying tasks that are far from being solved and highlighting those that are better addressed by the models. Since the landscape of tasks in IberBench is broad, we focus on task categories and we analyze the results across languages and LLMs, presenting the results per task category in Figure \ref{fig:all-performance-per-category}.

\begin{figure}[]
\centering
\includegraphics[scale=0.42]{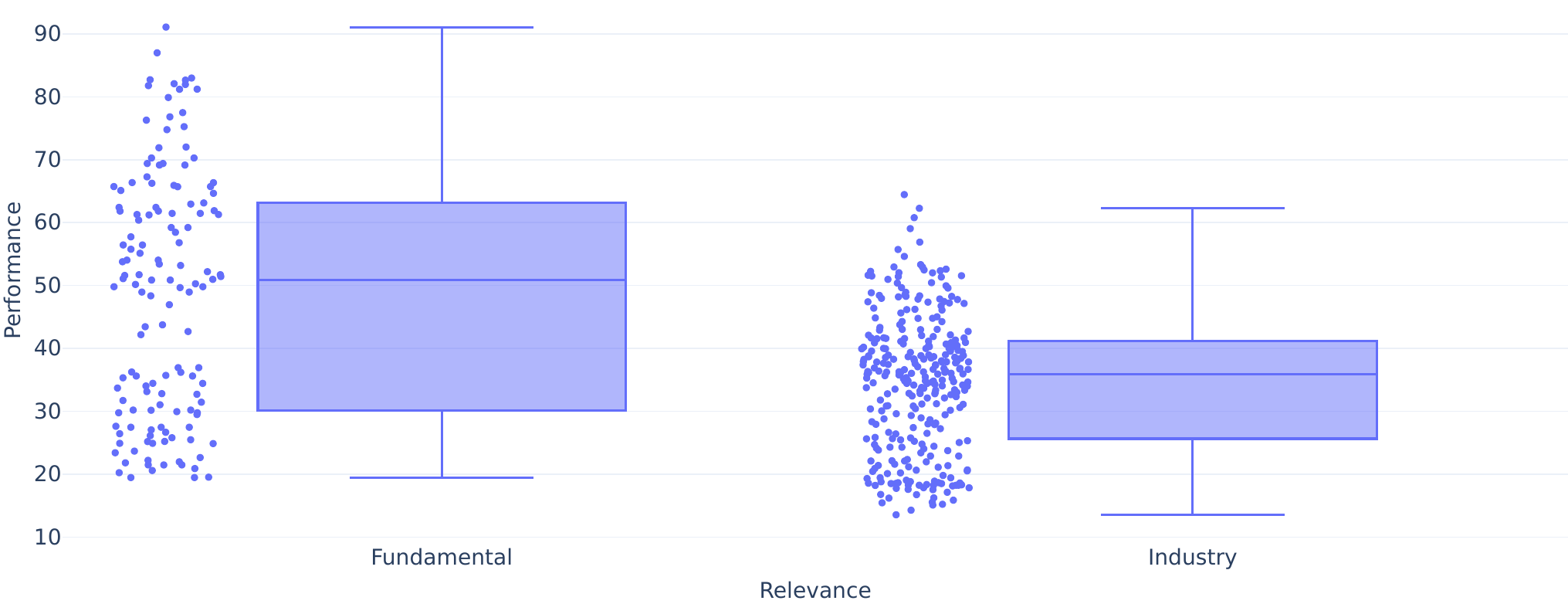}
\caption{\label{fig:all-performance-per-relevance} Performance in fundamental and industry-relevant tasks, averaged across all the languages and models.}
\end{figure}

\paragraph{LLMs perform better in fundamental tasks than on industry-relevant ones} The highest performances and medians are observed in Commonsense Reasoning, Question Answering, Reading Comprehension, Textual Entailment, and Paraphrase Detection. From Table \ref{tab:datasets-description-fundamental-relevant-task} we see that most of these tasks are fundamental, and dominate existing benchmarks \cite{baucells-etal-2025-iberobench}. In contrast, tasks with lower medians are more often industry-relevant, particularly Intent Classification, Stance Detection, Author Profiling, Summarization, and MGT Detection and Attribution. This performance gap is more clearly illustrated in Figure \ref{fig:all-performance-per-relevance}, which shows that models struggle more with industry-relevant tasks than with fundamental ones. This highlights the usefulness of IberBench, and suggests that benchmarks should shift their focus to other types of tasks to more realistically assess LLM capabilities.

\paragraph{Sequence-labeling tasks (for lexical analysis) remain challenging for LLMs} Results on Lexical Analysis, which involves a single task of detecting English borrowings in Spanish texts, are among the lowest across the benchmark, alongside those of Intent Classification. The best result in this task is 31.98\% Macro F$_1$, achieved by \texttt{phi-4}, while the other models do not exceed 20\%, with most scoring below 10\% (and half of these below 1\%). This suggests that (i) sequence-labeling tasks may not be well-suited for LLMs, and (ii) models with fewer than 14B parameters struggle to follow the instruction to generate text in the expected annotation schema. It remains uncertain whether larger models will be able to address the task effectively.

\paragraph{The random baseline is surpassed by far in fundamental tasks and it is hard to beat in some industry-relevant tasks} In 85.7\% of the fundamental tasks, the median results surpass the baseline, with some tasks, such as Reading Comprehension and Question Answering, showing the largest margins across all tasks. In industry-relevant tasks, the median surpasses the random baseline 60\% of the time, with a particularly large margin in Language Identification. Stance Detection, Fake News Detection, Author Profiling, Summarization, and MGT Detection and Attribution are the most challenging tasks. In MGT tasks, the baseline outperforms all the LLMs, highlighting the challenge for LLMs in identifying and attributing text generated by other LLMs.

\paragraph{Intent classification is hard... in Basque} Intent classification is a traditional, relatively simple task compared to more complex tasks such as summarization, mental health detection, or fake news detection. However, the LLMs' performances on this task are among the lowest in the benchmark. Notably, the Intent Classification category consists of only a single dataset in Basque. This suggests that most models struggle with Basque, a point we will further explore in our analysis per language. Despite this, even the best model achieves only around 20\% Macro F$_1$. Interestingly, this model is \texttt{sabia-7b}, an LLM further pre-trained for Portuguese.

\paragraph{What does it mean to ``break the ice''? LLM: It is a song by Britney Spears} Since LLMs can generate human-like text and handle complex classification tasks, one might expect it to accurately identify the language and demonstrate a high level of proficiency. This does not hold, as Language Identification, Proficiency Evaluation, and Linguistic Acceptability remain challenging for LLMs. This is particularly notable in the case of Proficiency Evaluation, which is framed as a question answering task requiring models to select the correct answer from proficiency exam questions. Despite this similarity, the median performance in Proficiency Evaluation is around 30\%, while the median performance in Question Answering tasks aimed to measure knowledge capabilities is around 50\%. We attribute these results to the specific challenges posed by the data. In Proficiency Evaluation, the category is dominated by Basque instances, a language that most models struggle with. For Language Identification, the difficulty likely stems from the domain, as all texts in this category are tweets: short, often informal messages that include slang, abbreviations, and non-standard writing styles.

\paragraph{Shared task participants still lead, but LLMs are narrowing the gap} Since all the tasks in IberBench originate from existing publications, we compare the LLMs' performance at the task category level with the best results reported in the literature. We focus only on task categories that include datasets from shared tasks. For this comparison, we average the performance of the best published models for the tasks within each task category and contrast them with the LLM results presented in Figure \ref{fig:all-performance-per-category}. In some tasks considered to be largely solved, such as Sentiment Analysis and Emotion Detection, the performance gap remains notable: the best LLM achieves 48.83\%, while the average of the best published results reaches 61.87\%. Large differences are also observed in Humor Detection (72.67\% vs. 84.26\%) and Lexical Analysis (31.98\% vs. 80.42\%). In other tasks, such as Fake News Detection (75.04\% vs. 76.66\%), Irony and Sarcasm Detection (62.92\% vs. 67.87\%), and Mental Health Detection (62.32\% vs. 68.76\%), LLMs still underperform compared to reported models, but the margins are smaller. It is important to note that our evaluation is conducted in a zero-shot setting, whereas the best published results are obtained with models fine-tuned on thousands of examples. We expect that under few- or many-shot conditions, the evaluated LLMs will outperform the best published results.

So far, we have discussed how well LLMs address the tasks in IberBench. However, we have not explicitly shown the ranking of LLMs per task category. For completeness, we present this ranking in Figure \ref{fig:ranking-per-task} of the Appendix.

\subsubsection{Performance per Iberian language}
\begin{figure}
\centering
\includegraphics[scale=0.48]{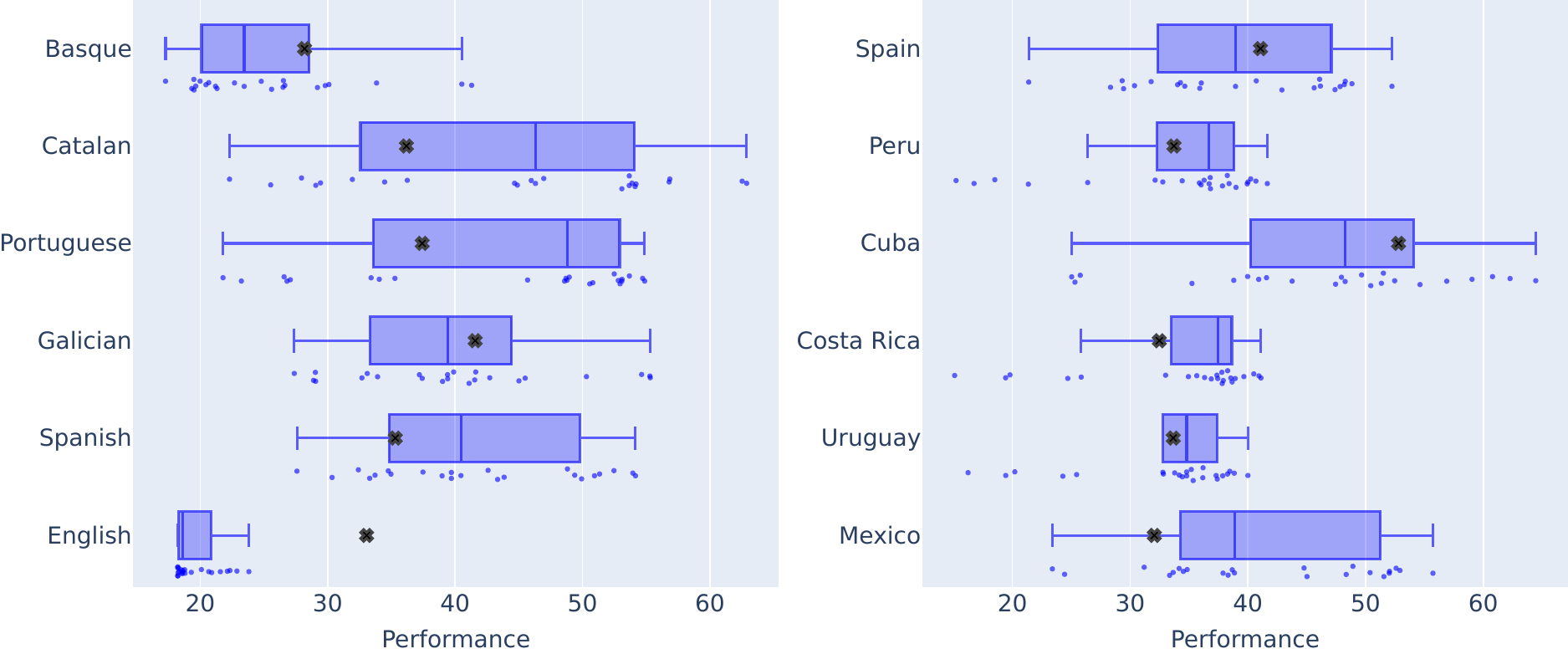}
\caption{\label{fig:all-performance-per-language} Performance per language and Spanish variety averaged across LLMs and tasks. For clarity, we remove the ``ambiguous'' Spanish variety.}
\end{figure}

\begin{figure}
\centering
\includegraphics[scale=0.42]{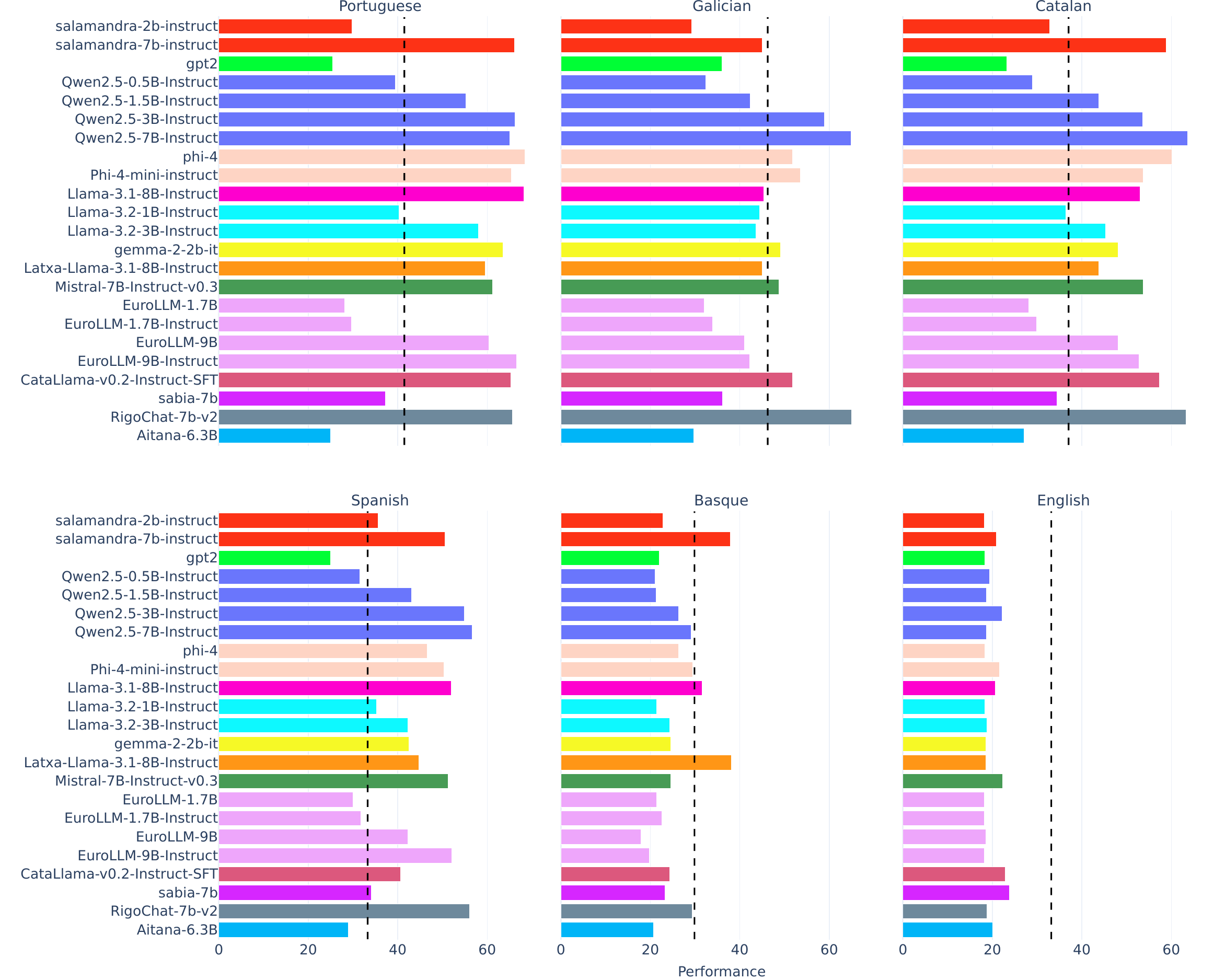}
\caption{\label{fig:all-performance-per-model-language} LLM performances in Iberian languages, averaged across tasks. Vertical lines denote the random baseline.}
\end{figure}

\begin{figure}
\centering
\includegraphics[scale=0.42]{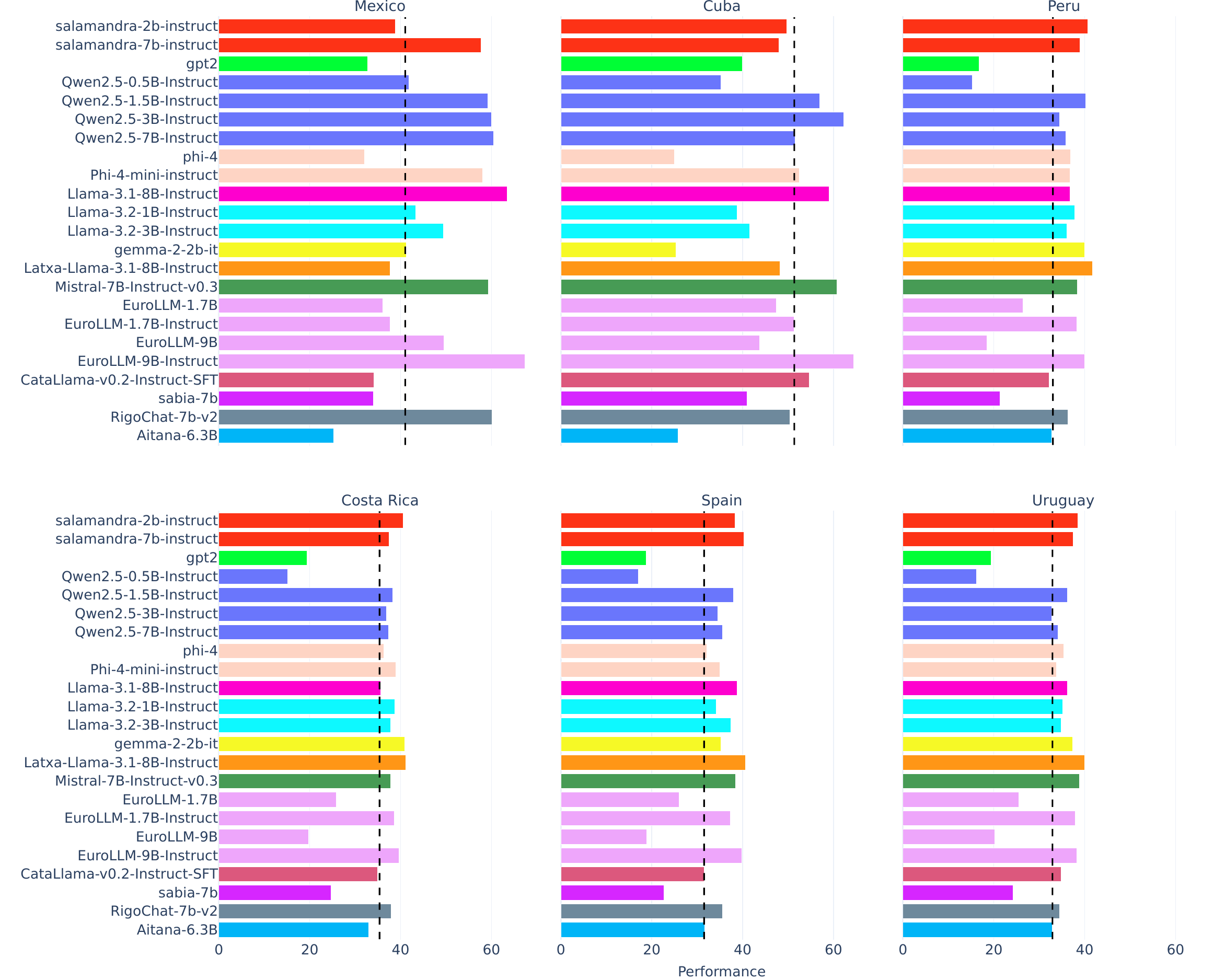}
\caption{\label{fig:all-performance-per-model-spanish-variety} Model performance in Spanish varieties. Vertical lines denote the random baseline.}
\end{figure}

We examine the performance of LLMs across Iberian languages, focusing on their challenges, distinctions, and unique characteristics, as well as the distribution of LLM performances across languages. We analyze (i) overall results per language averaged across LLMs and tasks (shown in Figure \ref{fig:all-performance-per-language}), and (ii) LLM results per language and Spanish variety averaged across tasks (shown in Figures \ref{fig:all-performance-per-model-language} and \ref{fig:all-performance-per-model-spanish-variety}, respectively).

\paragraph{English poses a unique challenge} Given that the datasets have been sourced mainly from evaluation campaigns with a focus on Iberian languages, the only category with English data is Machine Generated Text Detection and Attribution, which is very difficult for LLMs to solve without tailored techniques. As shown in Figure \ref{fig:all-performance-per-language}, a random baseline easily beats all the models. 

\paragraph{Galician and Basque present greater challenges than other languages} 
As shown in Figure \ref{fig:all-performance-per-model-language}, the performances in Catalan, Portuguese, and Spanish are remarkably better than the random baseline, with many models outperforming it by large margins. In contrast, while the top models in Galician outperform the random baseline by a large margin, only 34\% of models exceed it. In Basque, only three models barely beat the random baseline. This can be attributed to (i) the imbalance in task difficulty across languages, (ii) differences in language characteristics and (iii) differences in the amount of available resources for LLM training. Spanish, Portuguese, Catalan, and Galician are Romance languages that share many common features. However, despite recent efforts \cite{proyecto_aina, vladu2022proxecto}, the latter two languages, particularly Galician, are under-resourced in comparison. Galician is particularly interesting due to its similarity to Portuguese \cite{o2014galician}, meaning that language understanding and generation capabilities developed for Portuguese are expected to transfer to Galician as well. This is evident in the performance of \texttt{sabia-7b}, a base model further pre-trained for Portuguese. The performance of this model is very similar in both Portuguese and Galician, while it shows more variation in other languages.

\paragraph{Model rankings are remarkably consistent across languages and Spanish varieties} That is, a model $\mathcal{A}$ that outperforms another model $\mathcal{B}$ in a given language $\mathcal{L}_1$, it typically does so in another language $\mathcal{L}_2$ as well. 
To confirm this we compute Kendall's $\tau$ and Spearman's $\rho$ among the performance distributions for every pair of languages and varieties. In all cases, we find $\tau \approx \rho > 0.99$ with $p < 1\times10^{-11}$, underscoring the strong and statistically significant agreement in model rankings.

\paragraph{The curious case of Basque} From Figure \ref{fig:all-performance-per-model-language} we find that \texttt{Latxa-Llama-3.1-8B-Instruct}, a model obtained by further pre-training \texttt{Llama-3.1-8b-Instruct} with Basque data, performs similarly to \texttt{salamandra-7b-Instruct}, a multilingual LLM focused on Iberian languages. These three LLMs are the only ones that surpass the random baseline. Further pre-training for Basque improves performance in Basque, but this does not hold for Spanish, where \texttt{RigoChat-7b-v2} does not outperform \texttt{Qwen-2.5-7B-Instruct} in Spanish. Interestingly, further pre-training with a focus on Basque leads to catastrophic forgetting in other languages, as seen with \texttt{Latxa-Llama-3.1-8B-Instruct}: while it improves on the performance of \texttt{Llama-3.1-8B-Instruct} in Basque, it scores worse in every other language. In contrast, this effect is not observed when fine-tuning \texttt{Qwen-2.5-7B-Instruct} for Spanish (\texttt{RigoChat-7b-v2}). This highlights the distinctive characteristics of the Basque language in comparison to other Iberian languages: it lacks known linguistic relatives and is unrelated to the surrounding languages of the Iberian Peninsula \cite{b39047fe-1bd0-3e53-a7d1-21b44edfccbc}.

\paragraph{Two distinct types of Spanish varieties} We identify two groups of varieties in Figure \ref{fig:all-performance-per-language}, (i) varieties with low spread, lower median, and many outliers such as Peruvian, Costa Rican and Uruguayan, and (ii) varieties with higher spread, higher median, and no outliers like Cuban, Mexican and Spanish from Spain. From Figure \ref{fig:all-performance-per-model-spanish-variety} we see that some LLMs such as \texttt{EuroLLM-9B} or \texttt{Qwen-2.5-0.5b-Instruct} generally perform better in the latter group than in the former. Notably, Cuban and Spanish from Spain are the only varieties where the median does not surpass the baseline.

\paragraph{Multilingual models (including basque-tuned ones) outperform Spanish-specific models across Spanish varieties} As shown in Figure \ref{fig:all-performance-per-model-spanish-variety}, \texttt{RigoChat-7b-v2} is outperformed by a multilingual model in every Spanish variety. For instance, \texttt{Llama-3.1-8B-Instruct} surpasses it in Mexican and Cuban, while \texttt{salamandra-2b-Instruct} performs better in Uruguayan, Costa Rican, and Peruvian. Interestingly, \texttt{Latxa-Llama-3.1-8B-Instruct} ranks as the top-performing model in every Spanish variety except for Mexican and Cuban, often improving upon \texttt{Llama-3.1-8B-Instruct}. Despite showing signs of catastrophic forgetting in other languages, this improvement suggests that Latxa's fine-tuning may enhance either (i) generalization across Spanish varieties or (ii) performance on sentiment analysis. Notably, many of the Spanish varieties where Latxa excels contain only a single dataset focused on this task. From the model rankings per task category shown in Figure \ref{fig:ranking-per-task} of the Appendix we see that \texttt{Latxa-Llama-3.1-8B-Instruct} ranks second in Sentiment and Emotion Analysis, making the latter explanation more likely.

Akin to the ranking of LLMs per task from the previous section, we present the ranking of LLMs per Iberian language in Figure \ref{fig:ranking-per-language} of the Appendix.
\section{Conclusion} \label{sec:conclusion}
This work introduces IberBench, a multilingual and multitask benchmark designed to evaluate LLMs across Iberian languages in fundamental language skills and industry-relevant applications. 
Our benchmark focuses on the linguistic diversity of the Iberian Peninsula and Ibero-America, covering Spanish, Portuguese, Catalan, Basque, Galician, and English, as well as Spanish varieties such as Mexican, Uruguayan, Peruvian, Costa Rican, and Cuban. 
This scope enables more comprehensive and representative evaluations of LLMs in the underrepresented context of Iberian languages.

IberBench integrates 101 datasets compiled from evaluation campaigns and recent benchmarks, spanning 22 distinct task categories. 
To date, we have evaluated 23 LLMs ranging from 100 million to 14 billion parameters, covering models that are monolingual and multilingual, base models and models fine-tuned for chat applications, and more. 
We find that (i) LLMs struggle with industry-relevant tasks compared to fundamental language tasks, which dominate existing benchmarks, (ii) Galician and Basque present greater challenges than other languages, (iii) some tasks like lexical borrowing detection, intent classification, and machine-generated text detection remain largely unsolved, with top-performing LLMs barely surpassing a random guesser, and (iv) in other tasks such as sentiment analysis, humor detection, and fake news detection, LLMs are better than the random baseline but worse than most shared task submissions.

IberBench offers a complete open-source infrastructure for benchmarking, covering every stage from dataset processing and hosting to the incremental evaluation of LLMs. 
Evaluation results are integrated into a public leaderboard hosted on Hugging Face, providing easy access and fostering transparency. 
By open sourcing the IberBench pipeline, we hope to encourage community collaboration, reproducible research, and the responsible development of Iberian language technologies.

As future work, we aim to benchmark new LLMs on IberBench and progressively extend it with additional datasets as new evaluation campaigns with a focus on Iberian languages are launched within the NLP research community. 
\section*{Limitations and Ethical Considerations} \label{sec:limitations_ethical_considerations}
While IberBench provides a robust and flexible platform for evaluating language models in Iberian languages, it is important to highlight its limitations, especially related with data, modeling, evaluation, and computational resources.

The data that comprises IberBench was collected from existing LLM benchmarks, and shared tasks from NLP workshops. 
By gathering these datasets, IberBench includes a diverse set of Iberian languages, domains and tasks.
This also means that (i) IberBench is constrained to the data that is available from these sources with a permissive license, or where dataset authors have granted us explicit permission for use, (ii) it is primarily comprised of classification tasks, few generation tasks only including summarization, and one sequence labeling, and (ii) language varieties with a large number of speakers are underrepresented, as examples such as Argentinian Spanish or Brazilian Portuguese are not covered.
Moreover, some datasets may contain artifacts or inherent biases that can influence evaluation results. 
This can be likely (i) when labels or texts are obtained automatically, such as in author profiling or MGT detection tasks, (ii) where annotation is inherently difficult, like in mental health detection tasks, or (iii) it is subjective, e.g. in humor detection tasks.
Similarly to existing benchmarks, the task types and distributions across languages are imbalanced. 
All of these aspects can introduce bias to the evaluations and analyses, particularly when seeking fine-grained insights across languages or task categories.

The main modeling limitations stem from the prompting strategy. 
We employ a single prompt per task, designed according to established best practices. 
However, it is well known that LLMs are highly sensitive to the prompt phrasing and formatting. 
Alternative prompts could potentially yield better performance, particularly for sequence labeling tasks where the annotation schema may not align naturally with the format-following capabilities of an LLM.
Nevertheless, exploring a wide range of prompt variations is not feasible within our current experimental framework.
Another important limitation is our exclusive use of zero-shot evaluation\textemdash except for sequence-labeling tasks\textemdash which may underestimate model performance. 
Although few-shot prompting can improve results, its effectiveness is influenced by various factors and remains a topic of debate.
It is not yet widely adopted in practical applications due to data challenges, which motivates our choice of zero-shot evaluation to ensure consistency across tasks and better align IberBench with real-world usage scenarios.

The evaluation metrics used in IberBench also present drawbacks, particularly for assessing open-ended text generation. In such cases, we adopt ROUGE-1, following its use in other established benchmarks. However, ROUGE-1 focuses solely on unigram lexical overlap and does not capture other important dimensions of generation quality, such as fluency, coherence, hallucination, abstractiveness, or factual consistency.

The current evaluation process is restricted by the availability of computational and financial resources, which may bottleneck the frequency and scale at which new models can be assessed. 
Our first release of IberBench includes evaluations of LLMs with up to 14 billion parameters in 16-bit precision, excluding larger and closed-source LLMs. 

The deployment and use of benchmark evaluations in NLP entail several ethical considerations too. 
First, the selection of datasets and tasks can introduce biases if certain languages, varieties, or demographic groups are underrepresented. 
To mitigate this we strive to include a diverse range of tasks and data sources, although limitations in access and permissions currently hinder this goal.
Additionally, publishing model results and logs involves handling potentially sensitive data and outputs. 
We ensure that only publicly available and ethically sourced datasets are used in evaluations. 
We also avoid the inclusion of tasks that may contain harmful or discriminatory data unless they are explicitly designed to moderate this behavior.
\section*{Acknowledgements} \label{sec:acknowledgements}
We would like to express our sincere gratitude to the organizers of TASS, IberEVAL, IberLEF, and PAN workshops as well to the creators of existing LLM benchmarks in Iberian languages for providing access to the datasets included in this benchmark. The work from Symanto has been partially funded by the Instituto Valenciano de la Competitividad Empresarial (IVACE) under the grant IMINOK/2023/122.

\clearpage

\bibliographystyle{elsarticle-num} 
\bibliography{iberbench.bib}

\clearpage
\appendix
\section{Datasets and Sources}

Table \ref{tab:appendix-description-industry-relevant-task} lists the main URLs of the workshops, shared tasks and other sources we include in IberBench as industry-relevant tasks. Table \ref{tab:appendix-description-fundamental-relevant-task} shows the URLs of workshops, shared tasks and other sources that comprise the fundamental tasks of IberBench.

\begin{table}[h!]
\centering
\resizebox{1.00\textwidth}{!}{%
\footnotesize
\begin{tabular} {lllll}
\toprule
\textbf{Task} & \textbf{Subtask} & \textbf{Year} &  
\textbf{Language} & \textbf{URL} \\
\midrule
TweetLID & Language identification & 2014 & es & 
\url{http://komunitatea.elhuyar.eus/tweetlid/}\\
TASS & Emotion analysis & 2020 & es & \url{http://tass.sepln.org/2020/} \\
TASS & Sentiment analysis & 2020 & es-\{UY, MX, ES, CR, PE\} & \url{http://tass.sepln.org/2020/} \\
Author Profiling & Age detection & 2015 & es & \url{https://pan.webis.de/clef15/pan15-web/author-profiling} \\
Author Profiling & Gender detection &  2017 & es & \url{https://pan.webis.de/clef17/pan17-web/author-profiling}\\
MultiStanceCat & Stance Detection & 2018 & \{es, ca\} & \url{https://ceur-ws.org/Vol-2150/overview-Multistance18.pdf} \\
HAHA & Humor detection & 2019 & es &  \url{https://www.fing.edu.uy/inco/grupos/pln/haha/2019} \\
IroSvA & Irony detection & 2019 & es-\{CU, MX, ES\} & \url{https://ceur-ws.org/Vol-2421/IroSvA_overview.pdf}\\
MEX-A3T & Aggressiveness detection & 2019 & es-MX & \url{https://sites.google.com/view/mex-a3t2019} \\
DETOXIS & Aggressiveness detection & 2021 & es & \url{https://detoxisiberlef.wixsite.com/website} \\
DETOXIS & Improper language detection & 2021 & es &  \url{https://detoxisiberlef.wixsite.com/website} \\
DETOXIS & Insult detection & 2021 & es &  \url{https://detoxisiberlef.wixsite.com/website} \\
DETOXIS & Mockery detection & 2021 & es &  \url{https://detoxisiberlef.wixsite.com/website} \\
DETOXIS & Sarcasm detection & 2021 & es & \url{https://detoxisiberlef.wixsite.com/website} \\
DETOXIS & Toxicity detection & 2021 & es & \url{https://detoxisiberlef.wixsite.com/website} \\
EmoEvalES & Offensiveness detection & 2021 & es &   \url{https://competitions.codalab.org/competitions/28682} \\
EmoEvalES & Emotion analysis & 2021 & es & \url{https://competitions.codalab.org/competitions/28682}\\
EXIST & Sexism detection & 2021 & es & \url{https://nlp.uned.es/exist2021/} \\
EXIST & Sexism categorization & 2021 & es & \url{https://nlp.uned.es/exist2021/} \\
FakeDeS & Fake News detection & 2021 & es &   \url{https://sites.google.com/view/fakedes} \\
HAHA & Humor detection & 2021 & es & \url{https://www.fing.edu.uy/inco/grupos/pln/haha/} \\
MeOffendEs & Gender detection & 2021 & es &  \url{https://competitions.codalab.org/competitions/28679} \\
MeOffendEs & Offensiveness detection & 2021 & es & \url{https://competitions.codalab.org/competitions/28679} \\
Rest-Mex & Gender detection & 2021 & es-MX &   \url{https://sites.google.com/cicese.edu.mx/rest-mex-2021} \\
Rest-Mex & Sentiment analysis & 2021 & es-MX &  \url{https://sites.google.com/cicese.edu.mx/rest-mex-2021} \\
VaxxStance & Stance detection & 2021 & \{eu, es\} & \url{https://vaxxstance.github.io} \\
Par-Mex & Paraphrase detection & 2022 & es &   \url{https://sites.google.com/view/par-mex/home} \\
Rest-Mex & Sentiment analysis & 2022 & es-MX &  \url{https://sites.google.com/cicese.edu.mx/rest-mex-2022/home} \\
MentalRiskES & Eating disorder detection & 2023 & es &  \url{https://sites.google.com/view/mentalriskes} \\
MentalRiskES & Depression detection & 2023 & es & \url{https://sites.google.com/view/mentalriskes}\\
MentalRiskES & Depression categorization & 2023 & es & \url{https://sites.google.com/view/mentalriskes}  \\
HUHU & Fatphobia detection & 2023 & es &  \url{https://sites.google.com/view/huhuatiberlef23} \\
HUHU & Humor detection & 2023 & es &  \url{https://sites.google.com/view/huhuatiberlef23} \\
HUHU & LGBTIQ prejudice detection & 2023 & es & \url{https://sites.google.com/view/huhuatiberlef23} \\
HUHU & Racial prejudice detection & 2023 & es &  \url{https://sites.google.com/view/huhuatiberlef23} \\
HUHU & Women prejudice detection & 2023 & es & \url{https://sites.google.com/view/huhuatiberlef23} \\
DETESTS-Dis & Stereotype detection & 2024 & es & \url{https://detests-dis.github.io/} \\
IberAuTexTification & MGT detection & 2024 &\{en, es, eu, pt, ca, gl\} &  \url{https://sites.google.com/view/iberautextification} \\
IberAuTexTification & MGT attribution & 2024 & \{en, es, eu, pt, ca, gl\} & \url{https://sites.google.com/view/iberautextification}\\
PAWS & Paraphrase detection & 2019 & es &  \url{https://huggingface.co/datasets/google-research-datasets/paws-x} \\
&&& gl &  \url{https://huggingface.co/datasets/proxectonos/PAWS-gl} \\
&&& pt &  \url{https://huggingface.co/datasets/proxectonos/PAWS_pt} \\
&&& ca & \url{https://huggingface.co/datasets/projecte-aina/PAWS-ca} \\
XLSum & Text summarization & 2021 & \{es, pt\} &  \url{https://huggingface.co/datasets/csebuetnlp/xlsum} \\
BEC & Sentiment analysis & 2024 & eu & \url{https://huggingface.co/datasets/orai-nlp/basqueGLUE}\\
BHTC & Topic classification & 2024 & eu & \url{https://huggingface.co/datasets/orai-nlp/basqueGLUE} \\
caBreu & Text summarization & 2024 & ca & \url{https://huggingface.co/datasets/projecte-aina/caBreu} \\
ClinDiagnosES& Topic classification & 2024 & es & \url{https://huggingface.co/datasets/LenguajeNaturalAI/ClinDiagnosES} \\
FMTODeu & Intent classification & 2024 & eu & \url{https://huggingface.co/datasets/orai-nlp/basqueGLUE} \\
HateCheck & Hate speech detection & 2024 & pt & \url{https://huggingface.co/datasets/Paul/hatecheck-portuguese} \\
Parafraseja & Paraphrase detection & 2024 & ca & \url{https://huggingface.co/datasets/projecte-aina/Parafraseja} \\
\bottomrule
\end{tabular}
}
\caption{Industry-relevant tasks with URLs.} \label{tab:appendix-description-industry-relevant-task}
\end{table}

\begin{table}[h!]
\centering
\resizebox{1.00\textwidth}{!}{%
\footnotesize
\begin{tabular}{lllll}
\toprule
\textbf{Task} & \textbf{Subtask} & \textbf{Year} &  
\textbf{Language} & \textbf{URL} \\
\midrule
ADoBo & Lexical borrowing chunking & 2021 & es & \url{https://adobo-task.github.io/} \\
TE-ca & Textual entailment & 2021 & ca &  \url{https://huggingface.co/datasets/projecte-aina/teca} \\
OpenBookQA & Question answering & 2022 & es &  \url{https://huggingface.co/datasets/BSC-LT/openbookqa-es} \\
&&& ca &  \url{https://huggingface.co/datasets/projecte-aina/openbookqa_ca} \\
ARC & Question answering & 2024 & eu &  \url{https://huggingface.co/datasets/HiTZ/ARC-eu} \\
&&&  ca &  \url{https://huggingface.co/datasets/projecte-aina/arc_ca} \\
Belebele & Reading comprehension & 2024 & \{eu, es, pt, ca\} &  \url{https://huggingface.co/datasets/facebook/belebele} \\
CoLA & Linguistic acceptability & 2024 & es &  \url{https://huggingface.co/datasets/nbel/EsCoLA}\\
&&&  gl &  \url{https://huggingface.co/datasets/proxectonos/galcola} \\
&&& ca &  \url{https://huggingface.co/datasets/nbel/CatCoLA} \\
COPA & Commonsense reasoning & 2024 & eu &  \url{https://huggingface.co/datasets/HiTZ/XCOPA-eu} \\
&&& es & \url{https://huggingface.co/datasets/BSC-LT/COPA-es} \\
&&& ca &  \url{https://huggingface.co/datasets/projecte-aina/COPA-ca} \\
EusExams & Question answering & 2024 & \{eu, es\} &   \url{https://huggingface.co/datasets/HiTZ/EusExams}\\
EusProficiency & Proficiency evaluation & 2024 & eu & \url{https://huggingface.co/datasets/HiTZ/EusProficiency} \\
EusReading & Reading comprehension & 2024 & eu &  \url{https://huggingface.co/datasets/HiTZ/EusReading}\\
EusTrivia & Topic classification & 2024 & eu &  \url{https://huggingface.co/datasets/HiTZ/EusTrivia} \\ 
EusTrivia & Question answering & 2024 & eu & \url{https://huggingface.co/datasets/HiTZ/EusTrivia} \\
PIQA & Commonsense reasoning & 2024 & eu &  \url{https://huggingface.co/datasets/HiTZ/PIQA-eu} \\
TELEIA & Proficiency evaluation & 2024 & es & \url{https://huggingface.co/datasets/gonzmart/teleia} \\
QNLI & Textual entailment & 2024 & eu &   \url{https://huggingface.co/datasets/orai-nlp/basqueGLUE} \\
XNLI & Textual entailment & 2024 & es &  \url{https://huggingface.co/datasets/facebook/xnli} \\
&&& gl & \url{https://huggingface.co/datasets/proxectonos/xnli_gl} \\
&&& ca & \url{https://huggingface.co/datasets/projecte-aina/xnli-ca} \\
XStoryCloze & Question answering & 2024 & gl &  \url{https://huggingface.co/proxectonos/xstorycloze_gl} \\
&&& pt &  \url{https://huggingface.co/datasets/proxectonos/xstorycloze_pt} \\
&&& ca &  \url{https://huggingface.co/datasets/projecte-aina/xstorycloze_ca} \\
\bottomrule
\end{tabular}
}
\caption{Fundamental tasks with URLs.} \label{tab:appendix-description-fundamental-relevant-task}
\end{table}

\clearpage
\section{Model Rankings}

\begin{figure}[h!]
\centering
\includegraphics[scale=0.25]{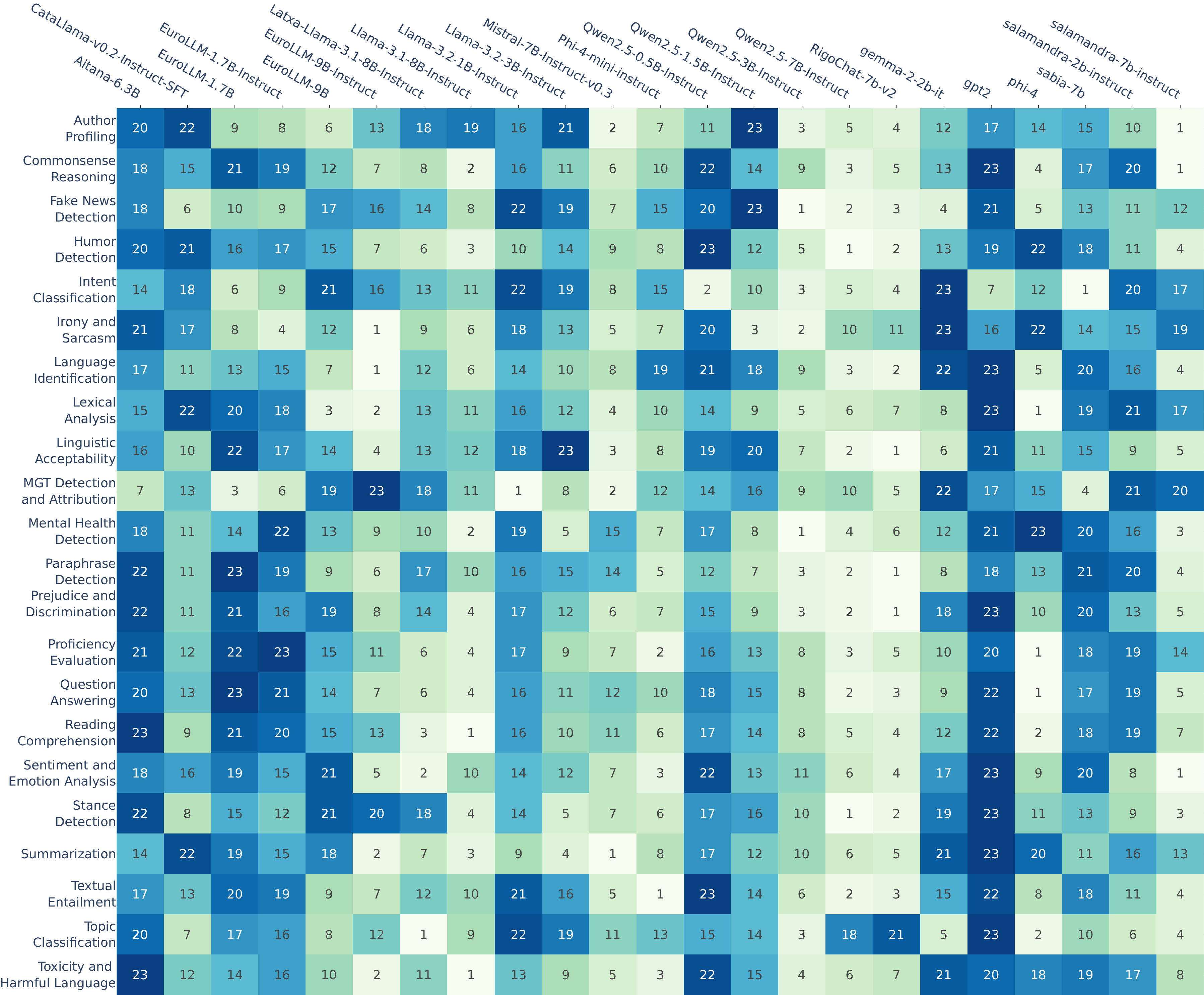}
\caption{\label{fig:ranking-per-task} Ranking of models per task category.}
\end{figure}

\begin{figure}[h!]
\centering
\includegraphics[scale=0.5]{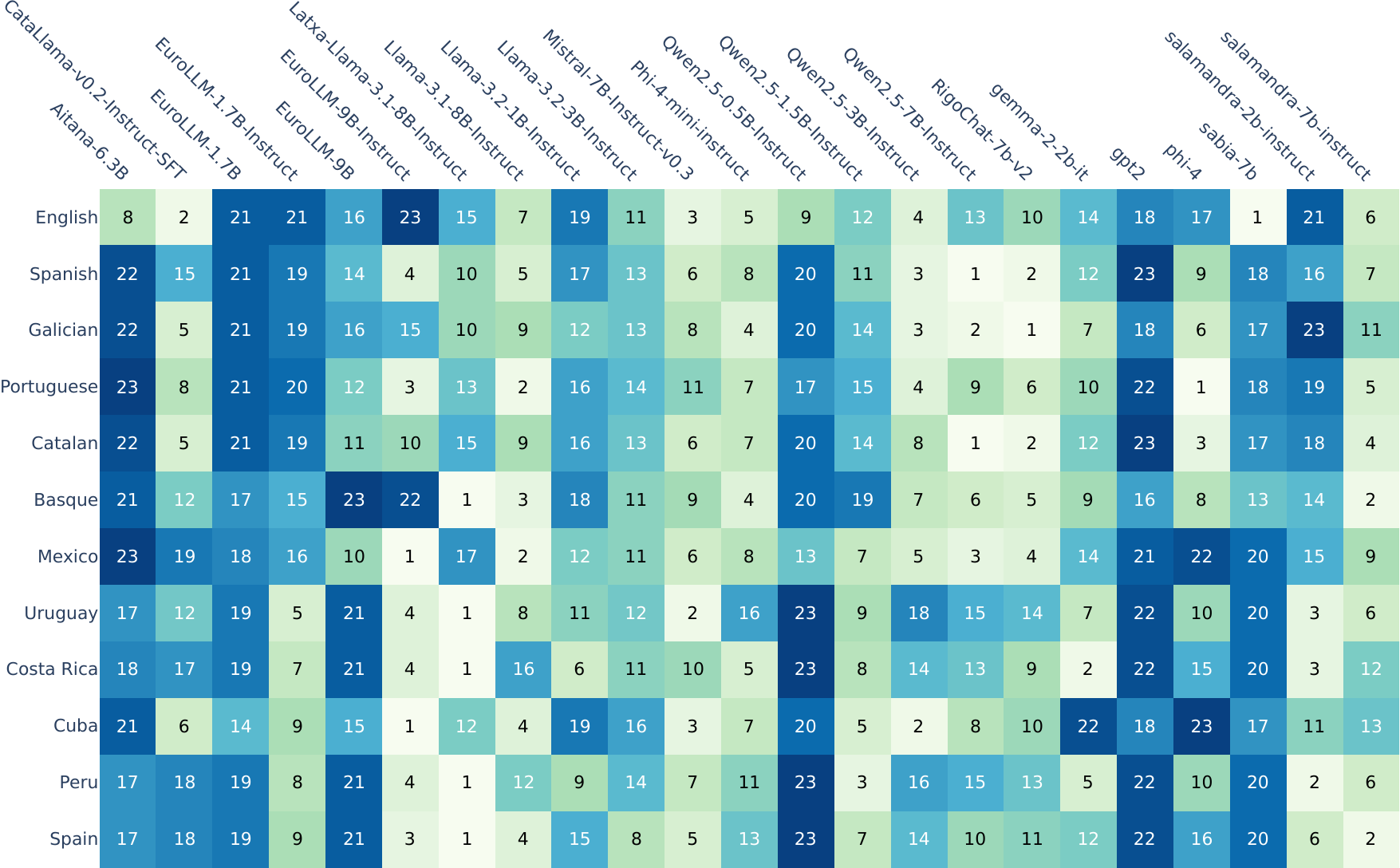}
\caption{\label{fig:ranking-per-language} Ranking of models per Iberian language.}
\end{figure}

Figures \ref{fig:ranking-per-task} and \ref{fig:ranking-per-language} show heatmaps of the rankings of each LLM in a given task and Iberian language respectively.

\end{document}